\documentclass[review]{clv2025}

\jvol{vv}
\jnum{nn}
\jyear{2026}

\dochead{Long Paper} 

\pageonefooter{Action editor: \{action editor name\}. Submission received: DD Month YYYY; revised version received: DD Month YYYY; accepted for publication: DD Month YYYY.}

\usepackage{amsmath}
\usepackage{booktabs}
\usepackage{multirow}
\usepackage{pifont}
\usepackage{subcaption}

\runningtitle{A Study of Crosslinguistic Influence in Language Models}
\runningauthor{Issam et al.}

\begin{document}

\title{A Study of Crosslinguistic Influence in Language Models}

\author{Abderrahmane Issam\thanks{Corresponding authors}, Yusuf Can Semerci, Jan Scholtes, Gerasimos Spanakis}

\affilblock{
    \affil{Maastricht University\\ \email{abderrahmane.issam@maastrichtuniversity.nl, y.semerci@maastrichtuniversity.nl, j.scholtes@maastrichtuniversity.nl, jerry.spanakis@maastrichtuniversity.nl}}
}

\maketitle

\begin{abstract}
The sequential acquisition of languages inevitably leads to Crosslinguistic Influence (CLI), where the syntactic properties of a first language (L1) impact the processing of a second language (L2). While modern language models exhibit robust cross-lingual transfer, the exact mechanisms governing how language dominance, relative proficiency, and typological distance dictate structural interference warrant deeper investigation. In this work, we systematically investigate CLI in artificial learners by training simultaneous and sequential bilingual models across 15 typologically diverse L1s and varying the Step of Exposure (SoE), defined as the specific training step at which the L2 is introduced. Utilizing crosslinguistic structural priming, we decouple latent CLI into distinct positive and negative transfer rates. Our evaluations reveal a critical computational tradeoff: while increased L1 dominance (higher SoE) strongly amplifies the correlation between structural transfer and syntactic distance, diminished L2 proficiency bottlenecks the model's capacity for positive transfer, leaving it highly vulnerable to persistent negative interference from distant L1s. Mechanistically, we demonstrate that L1 typological proximity physically dictates the cross-lingual overlap of L2 syntactic neurons. Furthermore, we uncover that explicit priming induces a dynamic layer-wise migration of grammatical resolution to the terminal layers of the network. Through targeted causal ablations, we confirm that deep-layer attention mechanisms exclusively drive this crosslinguistic transfer by routing the L1 structural prior into the final prediction. Ultimately, our findings demonstrate that CLI in language models is not an arbitrary artifact of capacity constraints, but a structured phenomenon fundamentally governed by the interplay of language dominance and proficiency.
\end{abstract}

\section{Introduction}

Most modern large language models (LLMs) are multilingual, trained jointly on data from various of languages \cite{grattafiori2024llama3herdmodels, openai2024gpt4technicalreport}. This training regime is typically dominated by a small number of high-resource languages, most notably English, alongside a long tail of lower-resource languages. The parameter sharing induced by this joint training gives rise to cross-lingual transfer, in which the dominant language's representations often benefit the acquisition and processing of lower-resource languages \cite{pires-etal-2019-multilingual, conneau-etal-2020-unsupervised, alkhamissi-etal-2025-llm, zhao2024how, wendler-etal-2024-llamas}. 

This transfer phenomenon is not unique to artificial learners: it has long been documented in human bilinguals, where it is studied under the heading of Crosslinguistic Influence (CLI) \cite{Hulk_Muller_2000, serratrice2013influence, vandijk2022, Siemund_2023}. Research on CLI seeks to understand how the representations of a bilingual's two languages interact and shape one another. Within psycholinguistics, structural priming has served as the central methodological tool for probing this interaction: exposure to a syntactic structure in a speaker's first language (L1) can influence both grammatical and ungrammatical productions in their second language (L2) \cite{hartsuiker2004, vandijk2023}. Furthermore, human studies indicate that the magnitude and direction of CLI are heavily modulated by the dynamic interplay of L1 dominance and L2 proficiency \cite{HARTSUIKER_BERNOLET_2017, van_dijk_2022}. Inspired by these cognitive frameworks, natural language processing (NLP) researchers have increasingly adopted cross-lingual structural priming to evaluate the development of shared representations in artificial language models \cite{arnett-etal-2025-acquisition, chang-etal-2024-multilinguality}.

While understanding multilingual representations has been extensively researched \cite{mueller-etal-2020-cross, deshpande-etal-2022-bert, chang-etal-2024-multilinguality, alastruey2025interferencematrixquantifyingcrosslingual, Doddapaneni2025}, studying CLI within massively multilingual architectures can obscure fundamental mechanisms. Vast data imbalances and the simultaneous joint training on hundreds of languages create a dense web of multidirectional interactions. Because numerous languages can simultaneously exert positive facilitation and negative interference on one another, isolating the exact transfer dynamics between any specific language pair becomes highly confounded. To isolate these dynamics, recent works have moved toward training controlled bilingual models that attempt to mimic human bilingual acquisition \cite{papadimitriou-jurafsky-2020-learning, oba-etal-2023-second, aoyama-schneider-2024-modeling}. Although these studies often demonstrate that L1 syntax affects the L2, yielding positive transfer for similar languages and negative transfer for distant ones, critical questions remain. Specifically, it is not clearly understood whether this effect is the result of genuine structural transfer or mere parameter interference. Moreover, the specific impact of co-activating L1 structures on L2 processing of grammatical and ungrammatical constructions has not been fully explored in these controlled settings. While studies of multilingual models have identified shared neurons across languages explaining cross-lingual transfer, their findings remain confounded and lack controlled ablations regarding language dominance. Furthermore, the relationship between structural co-activation, neuron overlap, and linguistic distance remains unexplored in models trained sequentially. 

Taken together, this literature leaves two central gaps. First, no work has examined how CLI and crosslinguistic priming interact with the dominance and proficiency effects established in human CLI literature. Second, although interpretability studies have identified candidate shared neurons, none have tested how these neurons behave under explicit structural co-activation, nor have they situated these analyses within controlled bilingual models immune to the confounds of massive multilinguality.

In this work, we address these gaps by using artificial language models as highly controlled testbeds in which we systematically ablate and investigate these variables. We train a suite of independently initialized simultaneous and sequential bilingual models, deliberately varying the Step of Exposure (SoE) to control the degree of L1 entrenchment and the resulting L2 proficiency. We evaluate English as the L2 alongside 15 typologically diverse L1s. To surface latent CLI, we employ crosslinguistic priming using direct L1 translations to activate established structural priors. Crucially, we move beyond raw accuracy by separately quantifying positive transfer (facilitation of grammatical structures) and negative transfer (priming of ungrammatical structures) on the BLiMP benchmark \cite{warstadt-etal-2020-blimp-benchmark}, and we track the layer-wise development of these CLI effects. Beyond these behavioral evaluations, we apply mechanistic interpretability techniques to trace the physical footprint of this transfer: we identify syntax-relevant L2 neurons and analyze their cross-lingual overlap, their layer-wise spatial distribution, and their component-level importance.

Our behavioral and mechanistic analyses show that sequential training establishes a structural prior that fundamentally constrains subsequent language acquisition. Our main contributions are as follows:

\begin{itemize}
    \item \textbf{Decoupling Dominance and Proficiency:} By varying the Step of Exposure, we demonstrate that L1 dominance and L2 proficiency jointly dictate CLI. Increased L1 entrenchment amplifies the correlation between linguistic distance and structural transfer. However, a critical computational tradeoff emerges: diminished L2 proficiency at higher SoE stages degrades the capacity for positive transfer through priming, while negative interference from the dominant L1 stubbornly persists (Section~\ref{sec:cli_effects}).
    \item \textbf{Bidirectional Priming and Target Susceptibility:} We explicitly ablate the direction of transfer and isolate the effect of language dominance. We demonstrate that while crosslinguistic influence is strictly bidirectional, its magnitude is profoundly asymmetric, proving significantly weaker when the target language is the dominant one (Section~\ref{sec:asymetry}).
    \item \textbf{Typological Proximity Dictates Neural Overlap:} We provide direct architectural evidence that L1 dominance physically shapes L2 parameter allocation. The cross-lingual overlap of identified L2 grammar neurons is modulated by linguistic distance, a typological consistency that strengthens significantly as L1 entrenchment increases (Section~\ref{fig:neurons_vs_distance}).
    \item \textbf{Priming Induces Deep-Layer Neural Migration:} We show that explicit crosslinguistic priming induces a dramatic spatial shift in syntactic processing. While unprimed L2 grammatical resolution peaks in the middle layers, explicit L1 priming forces active syntactic neurons to migrate into the terminal layers, dynamically shifting the computational locus of L2 resolution (Section~\ref{sec:neuron_layer}).
    \item \textbf{Attention Mechanisms Drive Context Integration:} Through targeted neural ablations, we isolate the causal driver of crosslinguistic transfer. We find that deep-layer attention units, rather than feed-forward (MLP) components, are responsible for modulating CLI. These mechanisms actively route the compressed L1 structural prior into the final target prediction, successfully leveraging shared pathways for similar languages while enforcing detrimental interference from distant ones (Section~\ref{sec:intervention_components}).
\end{itemize}

Ultimately, by adapting psycholinguistic paradigms to artificial learners, our findings demonstrate that crosslinguistic influence is not an arbitrary byproduct of model capacity constraints, but a structured effect shaped by factors such as L1 dominance and L2 proficiency, in a manner that closely parallels human CLI.

\section{Related Work}
\subsection{Crosslinguistic Influence}
The study of how two languages interact, known as Crosslinguistic Influence (CLI), remains a prominent research focus in both human bilingualism \cite{Hulk_Muller_2000, serratrice2013influence, vandijk2022, Siemund_2023} and artificial language learning \cite{papadimitriou-jurafsky-2020-learning, oba-etal-2023-second, yadavalli-etal-2023-slabert, alastruey2025interferencematrixquantifyingcrosslingual}. A primary mechanism for investigating CLI in humans is crosslinguistic priming. This paradigm examines how exposure to a syntactic structure in one language influences the processing or production of subsequent constructions in a target language \cite{Bock2003StructuralPA, hartsuiker2004, HARTSUIKER_BERNOLET_2017, Serratrice2022}. In the context of language models (LMs), crosslinguistic priming has served as evidence for the development of abstract structural representations \cite{sinclair2022, michaelov-etal-2023-structural, zhang-etal-2024-modeling-bilingual, arnett-etal-2025-acquisition}. To date, these LM studies have exclusively explored the priming of structures that are grammatical in both languages. However, the priming of structures that are ungrammatical in the target language is well documented in human psycholinguistic research \cite{vandijk2023, Baroncini_Torregrossa_2025}. Furthermore, while the impact of first language (L1) training on second language (L2) grammatical acquisition has been investigated \cite{papadimitriou-jurafsky-2020-learning, oba-etal-2023-second, yadavalli-etal-2023-slabert, constantinescu-etal-2025-investigating}, findings regarding the effect of L1 linguistic distance on L2 grammar remain mixed. This ambiguity highlights a need to understand how co-activating L1 structures via priming affects a model's preference for grammatical versus ungrammatical constructions in the L2. In this work, we evaluate the combined effects of L1 training and structural priming from linguistically distinct languages on an LM's ability to distinguish grammatical from ungrammatical sentences. We utilize direct translations into the L1 as primes to investigate the consequence of activating L1 structures on L2 processing. Translation-based crosslinguistic priming is an established phenomenon in human bilinguals \cite{Bangalore2016, Maier03082017}. Although we do not explicitly prompt the LMs to perform translation, we hypothesize that exposure to L1 translation equivalents will sufficiently activate L1 structures to induce measurable priming effects.

\subsection{Language Model Training for L2 Acquisition}

Prior work has extensively analyzed how multilingual training influences individual language acquisition \cite{mueller-etal-2020-cross, deshpande-etal-2022-bert, chang-etal-2024-multilinguality, alastruey2025interferencematrixquantifyingcrosslingual}. To mimic the dynamics of bilingual acquisition, some studies have trained models on unbalanced distributions of L1 and L2 data, or introduced the two languages sequentially (i.e., L1 followed by L2) to simulate late L2 exposure \cite{dhar-bisazza-2021-understanding, winther_2021, Roslund2022ModelingSP, papadimitriou-jurafsky-2020-learning, oba-etal-2023-second, aoyama-schneider-2024-modeling, constantinescu-etal-2025-investigating, arnett-etal-2025-acquisition}. While several of these studies artificially restrict training volumes to parallel human exposure environments, we argue that modeling robust structural transfer requires models to first achieve reliable grammatical competence. Because models trained strictly on human-scale data often exhibit sub-human performance \cite{oba-etal-2023-second, aoyama-schneider-2024-modeling}, we prioritize approaching human-like grammatical competence. This ensures that the observed CLI effects are representative of mature bilingual processing. Previous research using sequential training has investigated various effects on L2 acquisition, including syntactic transfer \cite{oba-etal-2023-second, yadavalli-etal-2023-slabert}, cognate facilitation \cite{winther_2021, Roslund2022ModelingSP}, and the prediction of L2 reading times \cite{aoyama-schneider-2024-modeling}. However, while the sequential introduction of an L2 is established in the literature, the specific impact of varying the step of exposure remains, to the best of our knowledge, unexplored. In this study, in addition to varying the step of exposure, we investigate its interplay with linguistic distance, orthographic script, and translation priming.

\subsection{Interpretability of Multilingual Syntax}
Understanding how multilingual models develop cross-lingual representations is a central question in natural language processing \cite{K2020Cross-Lingual, conneau-etal-2020-emerging, dufter-schutze-2020-identifying}. Interpretability studies have revealed that multilingual models allocate shared cross-lingual units \cite{varda-marelli-2023-data, mueller-etal-2022-causal}, overlapping circuits \cite{ferrando-costa-jussa-2024-similarity, zhang2024differentstructuralsimilaritiesdifferences}, and language-agnostic representations \cite{brinkmann-etal-2025-large, chang-etal-2022-geometry, libovicky-etal-2020-language}. Prior research demonstrates significant overlap among syntactic neurons across languages within these models \cite{mueller-etal-2022-causal, varda-marelli-2023-data, stanczak-etal-2022-neurons, antverg2022on}, with the degree of overlap correlating with linguistic distance \cite{kryvosheieva2026differenttypessyntacticagreement}. However, the activation of shared structures in independently trained models remains underexplored. In this context, \citet{zhang2024differentstructuralsimilaritiesdifferences} observed that independently trained monolingual models develop overlapping circuits across two languages for the task of Indirect Object Identification \cite{wang2023interpretability}, while \citet{conneau-etal-2020-emerging} demonstrated that monolingual models trained on different languages learn mutually aligned representations. Building on these findings, we investigate the mechanistic foundations of CLI and L1 priming. We identify L2 grammatical neurons in independently trained bilingual models and examine whether the L1 shapes L2 neural recruitment by correlating neuron overlap with linguistic distance. Finally, we ablate these identified neurons to evaluate their direct impact on both syntactic accuracy and priming effects.

\section{Bilingual Model Training}
\subsection{Language Selection}
\label{subsec:languages}

Choosing a representative and typologically diverse set of languages is critical for ensuring the generalizability of our findings. We select English as the target L2 language, as dictated by the availability of our evaluation datasets. To systematically investigate the dynamics of cross-lingual transfer and interference, we select a curated cohort of 15 L1 languages. This selection is governed by a careful balance of linguistic diversity, data availability, and computational feasibility. All chosen languages are classified as high-resource and are fully supported by our pretraining corpora and downstream processing tools.

The final language set spans multiple language families and writing systems. It includes eight languages that share the Latin script with English alongside seven that employ distinct scripts, such as Cyrillic, Arabic, and Devanagari. This specific design allows us to effectively isolate orthographic effects from purely syntactic phenomena. To properly model structural variance, the selected languages reflect a broad gradient of structural similarity to English. We quantify this variance using syntactic distance derived from the URIEL database language vectors (\texttt{syntax\_knn}) \cite{littell-etal-2017-uriel}, retrieved via the \texttt{lang2vec} tool\footnote{\url{https://github.com/antonisa/lang2vec}}. These vectors represent languages as binary arrays of typological properties, such as canonical word order and adposition placement, sourced from typological databases. To account for incomplete database coverage, the \texttt{\_knn} variant predicts any missing syntactic features using a $k$-nearest neighbors algorithm based on the known values of a language's phylogenetically and geographically closest neighbors. The final syntactic distance is formalized as: $ 1 - \text{sim}(\text{English}, X) $, where $\text{sim}$ represents the cosine similarity between the feature vectors of English and the target L1 language $X$. Table~\ref{tab:languages} details the complete list of L1 languages, ordered by increasing syntactic distance from English.
\begin{table}[h]
\centering
\small
\begin{tabular}{lccc}
\toprule
\textbf{L1 Language (ISO)} & \textbf{Family} & \textbf{Indo-European} & \textbf{Latin Script} \\ 
\midrule
German (de)   & Germanic      & \ding{51} & \ding{51} \\ 
Spanish (es)  & Romance       & \ding{51} & \ding{51} \\ 
French (fr)   & Romance       & \ding{51} & \ding{51} \\ 
Russian (ru)  & Slavic        & \ding{51} & \ding{55} \\ 
Polish (pl)   & Slavic        & \ding{51} & \ding{51} \\ 
Greek (el)    & Hellenic      & \ding{51} & \ding{55} \\ 
Indonesian (id)& Austronesian  & \ding{55} & \ding{51} \\ 
Finnish (fi)  & Uralic        & \ding{55} & \ding{51} \\ 
Vietnamese (vi)& Austroasiatic & \ding{55} & \ding{51} \\ 
Arabic (ar)   & Afroasiatic   & \ding{55} & \ding{55} \\ 
Thai (th)     & Kra-Dai       & \ding{55} & \ding{55} \\ 
Hindi (hi)    & Indo-Aryan    & \ding{51} & \ding{55} \\ 
Korean (ko)   & Koreanic      & \ding{55} & \ding{55} \\ 
Turkish (tr)  & Turkic        & \ding{55} & \ding{51} \\ 
Japanese (ja) & Japonic       & \ding{55} & \ding{55} \\ 
\bottomrule
\end{tabular}
\caption{Typological classification and script overlap of the 15 selected L1 languages relative to English (L2), ordered by increasing syntactic distance.}
\label{tab:languages}
\end{table}

\subsection{Data Acquisition and Code-Switching Mitigation}
\label{subsec:data_curation}

To construct a clean pretraining corpus for each target L1 language, we extract documents from the FineWeb \cite{penedo2024the} web-scraped dataset. Specifically, we stream the 10-billion token sample (\texttt{sample-10BT}) for English and the language-specific splits from FineWeb-v2 \cite{penedo2025fineweb2pipelinescale} for the L1 languages. Given that web data exhibits a high frequency of code-switching, which is known to enhance cross-lingual transfer \cite{ijcai2020p533, wang-etal-2025-investigating-scaling}, we implement a rigorous filtering pipeline to minimize it. Our primary objective is to isolate and understand the fundamental interplay between linguistic similarity and crosslinguistic influence, rather than conflating these structural dynamics with varying degrees of English code-switching across different language subsets.

The filtering protocol applies a coarse-to-fine sequence of language identification and code-mixing mitigation. First, we employ a fast sentence-level pre-filter using FastText \cite{joulin-etal-2017-bag} to discard sentences where the predicted target language confidence falls below a threshold of 0.70. Second, because sentence-level classifiers often fail to detect fine-grained code-switching or loanword insertion, we compute a token-level Code-Mixing Index (CMI) for all surviving sentences. To maximize token classification accuracy across diverse scripts, we implement a hybrid detection strategy: non-Latin tokens are mapped deterministically to their respective languages using Unicode script properties via \texttt{unicodedata}, whereas Latin-script tokens longer than three characters are resolved using a statistical language detector, \texttt{Lingua-py}\footnote{\url{https://github.com/pemistahl/lingua-py}}. To eliminate spurious noise from ambiguous or short tokens, predictions with a confidence lower than 0.60 default to the dominant language of the corpus.

Formally, for a given sentence containing $N$ valid tokens, let $w_i$ denote the language assigned to token $i$. The CMI for the sentence is defined as follows:

\begin{equation}
\text{CMI} = 100 \times \left(1 - \frac{\max_{L} \sum_{i=1}^N \mathbb{I}(w_i = L)}{N}\right)
\end{equation}
where $\max_{L}$ is the count of the most frequent language in the sentence. Sentences yielding a $\text{CMI} > 5.0$ are classified as code-switched and are subsequently removed. This pipeline continues until a uniform target of exactly 64M sentences is successfully collected for each language.

Mitigating code-switching remains an open challenge in NLP \cite{burchell-etal-2024-code}, particularly when balancing precision with the computational efficiency required to filter millions of sentences. Our pipeline represents a systematic effort to address this confounding factor, which has been largely overlooked in prior cross-lingual transfer literature despite its critical implications. While our approach remains inherently bounded by the baseline accuracy of underlying language identification tools, it substantially reduces code-switching density. The impact of this pipeline is highly pronounced in practice; for instance, on the Indonesian subset, our approach identifies and filters approximately 48M code-switched sentences. The result of this process is 64M lines of monolingual text data per language.

\subsection{Bilingual Model Training}
\label{subsec:bilingual_training}

\begin{figure}[t!]
    \includegraphics[width=0.45\textwidth]{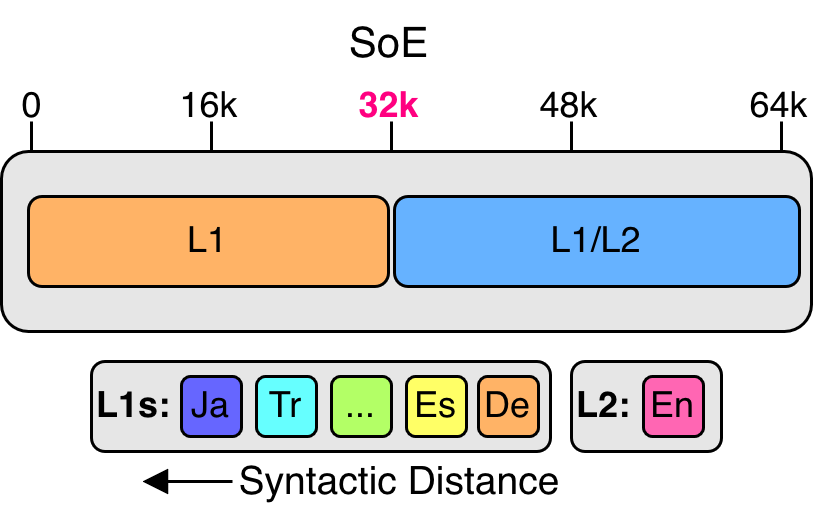}
    \caption{Setup for modeling CLI using LMs. We manipulate the Step of Exposure (SoE) by introducing the L2 at specific training steps after pretraining on the L1. The second stage of training uses interleaved L1/L2 sequences to maintain L1 influence. We select 15 L1 languages with increasing syntactic distance from the L2.}
    \label{fig:intro_aeo}
\end{figure}

Following prior work \cite{chang_word2022, arnett-etal-2025-acquisition}, we train bilingual GPT-2 models to investigate how varying the Step of Exposure (SoE) to a second language impacts crosslinguistic influence (CLI). In human bilinguals, L1 dominance and L2 proficiency are known to strongly modulate CLI and priming effects \cite{vandijk2022, HARTSUIKER_BERNOLET_2017}. To simulate this dynamic, we define SoE as the specific training step at which L2 data is introduced. As illustrated in Figure \ref{fig:intro_aeo}, increasing the SoE results in longer initial L1 training. This entrenches L1 representations and simultaneously reduces the total volume of L2 input, thereby increasing L1 dominance and decreasing L2 proficiency. Once the L2 is introduced, we maintain exposure to the L1. This strategy prevents catastrophic forgetting \cite{constantinescu-etal-2025-investigating, arnett-etal-2025-acquisition}, compensating for the critical period effects that generally protect human learners from such forgetting \cite{Lenneberg01121967, bylund2019}. While this framework simplifies human language acquisition, it provides a principled method to establish a clear language dominance hierarchy.

To prepare the training data, we first train a SentencePiece tokenizer \cite{kudo-richardson-2018-sentencepiece} for each language pair. To prevent vocabulary bias, we diverge from \cite{arnett-etal-2025-acquisition} by training the tokenizer on a strictly balanced dataset of 10M lines, comprising 5M lines from each language. The corpus is then processed into shuffled sequences of 256 tokens. For each language, we sample one billion tokens for training and one million tokens for evaluation.

All models undergo a total of 64k training steps. During monolingual training phases, batches consist entirely of sequences from a single language, while bilingual training phases utilize batches with interleaved L1 and L2 sequences. We systematically vary the SoE by initiating bilingual training at step 0, 16k, 32k, and 48k. Standard language model training typically employs a linear learning rate scheduler. However, a decaying learning rate would confound our results, as later SoE conditions would process L2 data at a substantially lower learning rate than the L1 data. To isolate the effect of SoE from optimization dynamics, we utilize a constant learning rate following a 5,000-step warmup. The isolated impact of the learning rate scheduler is further explored in our ablation study (Section \ref{sec:ablate_linear}).

For every bilingual pair, we train a corresponding monolingual baseline model using the identical tokenizer. This baseline is trained exclusively on the L2 for the full 64k steps, simulating a monolingual speaker of the same chronological age as the bilingual models. Detailed training hyperparameters are provided in Appendix \ref{sec:training_details}.

\vspace{-0.2cm}

\section{CLI Analysis}
\subsection{Evaluating Transfer Effects}
\label{sec:method_transfer_effects}

To quantify the effect of the L1 on L2 grammatical knowledge, we utilize the BLiMP benchmark \cite{warstadt-etal-2020-blimp-benchmark}, which consists of 67,000 minimal pairs of acceptable and unacceptable sentences across 67 linguistic categories. We evaluate model performance based on sentence-level surprisal. For a given pair $(s_\text{acc}, s_\text{unacc})$, the model is considered correct if it assigns a higher pseudo-log-likelihood to the acceptable sentence:
$$ \log P(s_\text{acc}) > \log P(s_\text{unacc}) $$

To account for variations in sequence length, we compute the length-normalized pseudo-log-likelihood as:
$$ \log P(s) = \frac{1}{N} \sum_{t = 1}^{N} \log P(x_t \mid x_{< t}) $$
where $N$ represents the sequence length.

Prior studies have primarily focused on overall accuracy when evaluating the impact of the L1. In contrast, we conduct a fine-grained analysis by examining positive ($\text{CLI}^{+}$) and negative ($\text{CLI}^{-}$) transfer separately relative to the equivalent monolingual baseline. Let $\mathcal{C}_{\text{mono}}$ and $\mathcal{C}_{\text{bi}}$ represent the sets of instances correctly predicted by the monolingual and bilingual models, respectively, within an evaluation dataset of size $N$. We define positive transfer as the proportion of instances that the monolingual model predicts incorrectly but the bilingual model predicts correctly, and negative transfer as the converse:
\begin{equation}
\text{CLI}^{+} = \frac{|\mathcal{C}_{\text{bi}} \setminus \mathcal{C}_{\text{mono}}|}{N}, \quad \text{CLI}^{-} = \frac{|\mathcal{C}_{\text{mono}} \setminus \mathcal{C}_{\text{bi}}|}{N}
\end{equation}

This approach provides a granular perspective on transfer effects that aggregate accuracy metrics inherently obscure. For instance, if an L1 induces significant but equal amounts of positive and negative transfer, the net change in accuracy would be zero, effectively masking the underlying crosslinguistic influence. Furthermore, by comparing the bilingual model against a monolingual baseline trained with an identical tokenizer, this methodology successfully isolates transfer effects from vocabulary artifacts.

\subsection{Crosslinguistic Priming}
To investigate the influence of shared or connected representations on L2 processing, we employ a crosslinguistic priming paradigm inspired by psycholinguistic studies \cite{HARTSUIKER_BERNOLET_2017, vandijk2023}. Following the approach of \citet{arnett-etal-2025-acquisition}, we evaluate whether the presence of an L1 prime affects the model's preference for L2 grammatical vs. ungrammatical structures. Since BLiMP benchmark lacks parallel translations, we generate L1 primes by translating the acceptable sentence of each minimal pair into the corresponding L1 using the nllb-200-3.3B model\footnote{\url{https://huggingface.co/facebook/nllb-200-3.3B}}. We note that all the L1 languages in our study are in the high resource category of this model ensuring high quality translations \cite{nllbteam2022languageleftbehindscaling}. To measure the priming effect, we prepend the L1 translation as context and calculate the length-normalized negative log-likelihood (surprisal) of the subsequent L2 acceptable and unacceptable sentences. Formally, we evaluate the conditional log-probability: $\log P(X | \text{Prime})$, where $X \in {s_\text{acc}, s_\text{unacc}}$ and the Prime is the L1 translation of $s_\text{acc}$. By comparing $CLI_{L1 \to L2}$ before and after priming, we quantify the degree to which L1 structures facilitate or interfere with L2 processing. Note that this is not applied to the monolingual models.
\subsection{Finding Grammar Neurons}
\label{sec:finding_neurons}

To investigate whether mechanistic evidence exists for crosslinguistic influence, we complement our behavioral evaluation on the BLiMP benchmark with an internal analysis of the L2 representations in each bilingual model. Specifically, we aim to identify specialized grammar units and observe how they are modulated by the L1. We draw inspiration from functional localization techniques used to isolate language-sensitive regions in the human brain based on their selective responses to specific stimuli. By applying this neuroscientific framework to language models, we adopt a unit identification strategy similar to recent approaches in mechanistic interpretability \cite{alkhamissi-etal-2025-llm, kryvosheieva2026differenttypessyntacticagreement}.

Prior work \cite{kryvosheieva2026differenttypessyntacticagreement} localized multilingual grammar units by analyzing the post-residual outputs of entire transformer blocks. This macroscopic view inherently obscures the distinct mechanisms at play within the individual feed-forward and attention sub-layers. To achieve a more granular understanding, we extend this methodology by independently analyzing the internal Multi-Layer Perceptron (MLP) and Attention components (i.e., query, key, value, and output projections). Rather than measuring the output after the residual connection, we isolate the intermediate representations within the MLP network, which function as classical feed-forward neurons. To compute token-wise impact scores for each attention component, we adapt the formulation introduced by \cite{zhao2024how}, referring the reader to their work for full technical details.

Our localization procedure isolates units that selectively and robustly respond to valid grammatical structures. Using the BLiMP dataset, we define acceptable sentences as our positive condition and their unacceptable minimal pairs as the negative control. Similar to \cite{kryvosheieva2026differenttypessyntacticagreement}, we conduct this unit identification process independently for each linguistic category within BLiMP. For every targeted module, we extract the absolute magnitude of the last-token activations across both conditions. Following the statistical approach of \cite{alkhamissi-etal-2025-llm}, we perform an independent Welch's t-test between the positive and negative activation distributions for each individual unit. Grammar units are then formally identified by selecting the top 5\% of neurons within each module that yield the highest t-statistic, isolating those that significantly and reliably activate for acceptable grammatical sequences over unacceptable ones.

\section{Results and Analysis}
\subsection{CLI Effects on BLiMP}
\label{sec:cli_effects}
\subsubsection{Correlation of CLI and Linguistic Distance}
\label{sec:cli_distance}
\begin{figure*}[t!]
    \centering
    \includegraphics[width=1.\textwidth]{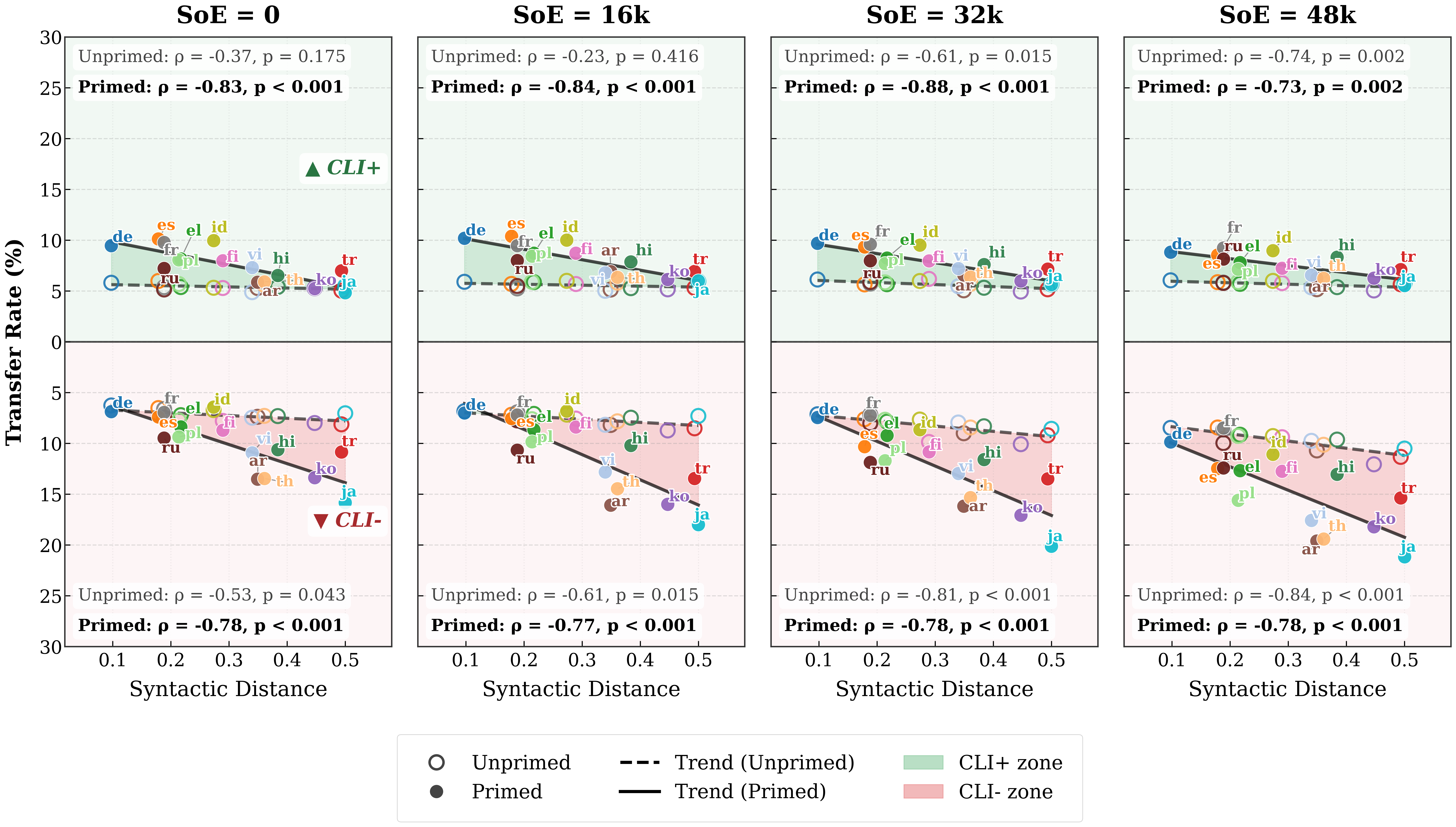}
    \caption{Correlation of positive and negative transfer rates with syntactic distance across four Step of Exposure (SoE) checkpoints. The upper region (green background) displays positive transfer rates ($\text{CLI}^+$), while the lower region (red background) illustrates negative transfer ($\text{CLI}^-$). Filled markers and solid trendlines represent the explicitly primed condition, demonstrating higher absolute magnitudes of transfer. Hollow markers and dashed trendlines denote the unprimed condition. Spearman rank correlations ($\rho$) and $p$-values are reported for each panel, highlighting how the rank-order relationship between syntactic distance and transfer strengthens as SoE increases.}
    \label{fig:prime_effect}
\end{figure*}

To isolate the specific mechanisms of crosslinguistic influence, we evaluate positive and negative CLI independently, following the established methodology. As illustrated in Figure \ref{fig:prime_effect}, plotting these distinct transfer rates against syntactic distance yields a clear structural hierarchy across all four Step of Exposure (SoE) checkpoints. A primary observation is that explicit translation priming significantly enhances the correlation between syntactic distance and both positive and negative transfer effects.

In the unprimed condition, the absolute magnitude of transfer is substantially lower and exhibits minimal variance across languages, resulting in a visually shallow trendline. However, as the SoE increases, effectively increasing L1 dominance by entrenching L1 representations prior to L2 introduction, the unprimed trendline steepens considerably. For negative transfer, the slope steepens from $\beta = -2.68$ at SoE=0 to $\beta = -7.07$ at SoE=48k. Crucially, the rank-order relationship between syntactic distance and transfer becomes highly consistent as training progresses. Even at lower absolute transfer volumes, syntactic distance reliably predicts the hierarchy of negative transfer. The Spearman correlation for unprimed negative transfer strengthens significantly from a marginal $\rho = -0.37$ ($p = 0.175$) at SoE=0 to a highly robust $\rho = -0.74$ ($p = 0.002$) at SoE=48k. This indicates that sequentially driven L1 dominance leads to transfer effects strongly modulated by structural proximity.

Finally, because our experimental design relies on automatic translation to generate the primes, it is crucial to ensure that the observed transfer effects are driven by genuine crosslinguistic influence rather than the translation quality or idiosyncratic behaviors of a single translation model. To rule out this confounder, we replicate the prime translations using the MADLAD\footnote{\url{https://huggingface.co/google/madlad400-3b-mt}} system, selecting a variant of similar capacity ($\sim$3B parameters) to our primary NLLB model. We compute the Spearman rank correlation of the log-probability differences between acceptable and unacceptable BLiMP sentences across the primed conditions generated by both translation systems. The resulting correlation is exceptionally strong across all languages and SoE checkpoints, ranging from a minimum of $\rho = 0.923$ (Hindi) to a maximum of $\rho = 0.981$ (Vietnamese), with an overall mean of $\rho \approx 0.96$ ($p < 0.001$ in all conditions). This robust consistency verifies that the measured structural transfer is an intrinsic property of the bilingual model's representation space, successfully demonstrating that our findings are not an artifact of the automatic translation system.

\subsubsection{Per-Layer CLI Effects}
\label{sec:layer_cli}
To understand exactly where the crosslinguistic influence from translation priming originates within the network, we apply the logit lens \cite{nostalgebraist2020logitlens} technique to probe the model's intermediate representations. By projecting the hidden states of intermediate layers directly into the vocabulary space using the unembedding matrix, we calculate the intermediate log-probabilities and evaluate the positive and negative transfer rates at each transformer block, exactly as outlined in Section~\ref{sec:method_transfer_effects}. To isolate the specific impact of the prime, we compute the difference between the transfer rates of the primed and unprimed models ($\Delta$ Transfer Rate).

\begin{figure*}[t!]
    \centering
    \includegraphics[width=1.\textwidth]{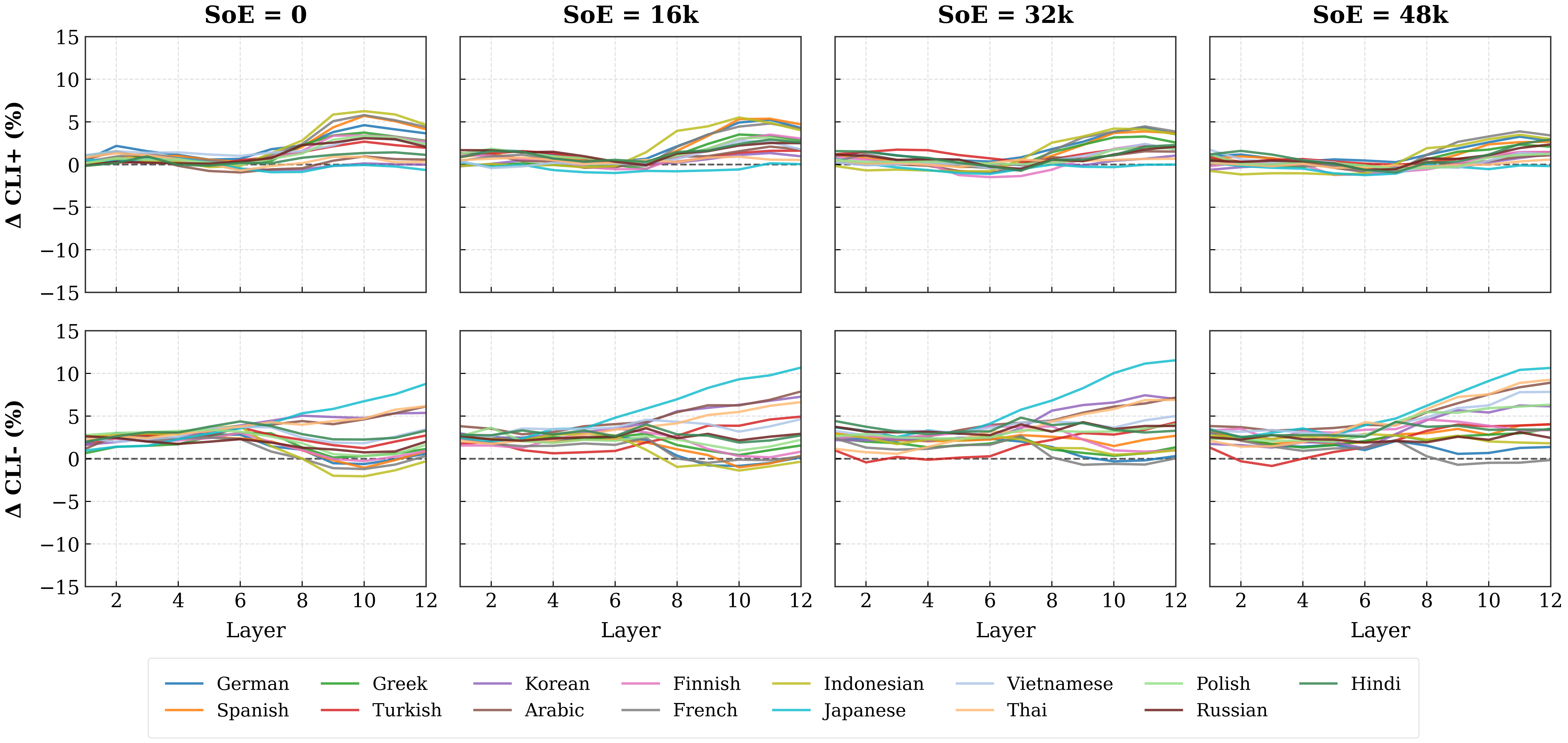}
    \caption{Layer-wise $\Delta \text{CLI}^+$ (top) and $\Delta \text{CLI}^-$ (bottom) for each L1 at four Step of Exposure (SoE) checkpoints. Rates are computed as the difference between primed and unprimed models.}
    \label{fig:layer_prime_effect}
\end{figure*}

Figure~\ref{fig:layer_prime_effect} illustrates these layer-wise trajectories. In the early layers (layers 1 through 6), the differential positive transfer remains near zero across all languages. The differential negative transfer, by contrast, starts at a baseline strictly greater than zero across all priming conditions. This indicates a universal early-layer interference that occurs regardless of the typological distance of the L1. 

A clear divergence emerges after layer 6. For languages that are typologically close to English (German, French, Spanish), negative transfer steadily declines while positive transfer sharply rises. Conversely, for typologically distant languages (Japanese, Korean, Thai), negative transfer climbs continuously while positive transfer remains suppressed. The structural overlap between an L1 and English thus dictates the trajectory of crosslinguistic influence exclusively in the deeper layers.

\subsubsection{Layer-Wise Dynamics of Crosslinguistic Influence}
\label{sec:dynamics_cli}
To provide a more granular perspective on these representational shifts, we move beyond binary accuracy metrics and analyze crosslinguistic influence continuously via surprisal differences. By calculating the difference in surprisal between acceptable and unacceptable sentences, we capture both the magnitude and direction of the grammatical preference. We then compute the cosine similarity of these surprisal-difference vectors between every possible language pair across all layers. We hypothesize that as the linguistic distance between two L1s increases, the cosine similarity of their grammatical transfer footprints should decrease, yielding a negative correlation.

\begin{figure*}[t!]
    \centering
    \includegraphics[width=1.\textwidth]{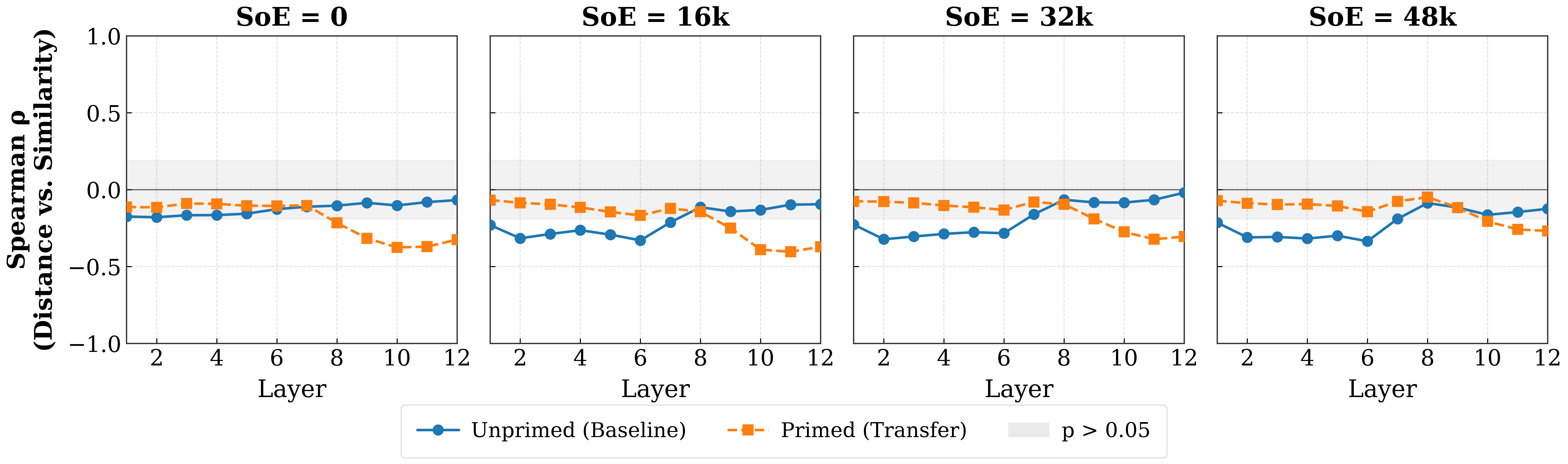}
    \caption{Layer-wise Spearman correlation ($\rho$) between linguistic distance and the cosine similarity of surprisal differences.}
    \label{fig:surprisal_diff_corr}
\end{figure*}

Figure~\ref{fig:surprisal_diff_corr} plots the layer-wise Spearman correlation ($\rho$) between structural distance and surprisal cosine similarity. In the unprimed models (with the exception of SoE=0), the early layers exhibit a statistically significant negative correlation. This indicates that latent L1 biases learned during pretraining actively shape early L2 representations. However, this correlation weakens steadily into the non-significant regime ($p > 0.05$) in the upper layers, suggesting the model progressively suppresses interfering L1 representations to optimize for monolingual English output.

The primed models exhibit the inverse pattern, providing a mechanistic explanation for the layer-wise divergence observed in Figure~\ref{fig:layer_prime_effect}. In the early layers, the correlation is not statistically significant. Because the L1 prime is prepended to the L2 target, early layers process them largely in isolation due to their reliance on shallow, localized processing \cite{jawahar-etal-2019-bert, tenney-etal-2019-bert}. It is not until layer 8 that the primed models show a statistically significant negative correlation. This delayed onset aligns with recent findings on in-context learning, where early layers compress contextual cues into an abstract subspace while later layers utilize this structure for inference \cite{jiang2026from, xu2026emergent}. In our setting, this transition point marks where the network actively integrates the compressed L1 context, causing crosslinguistic influence to finally manifest based on structural similarity. Together, these results demonstrate that the L1 prime acts as a structural prior, modulating L2 predictions only after high-level context is integrated in the deeper layers.

\begin{figure*}[t!]
    \centering
    \includegraphics[width=1.\textwidth]{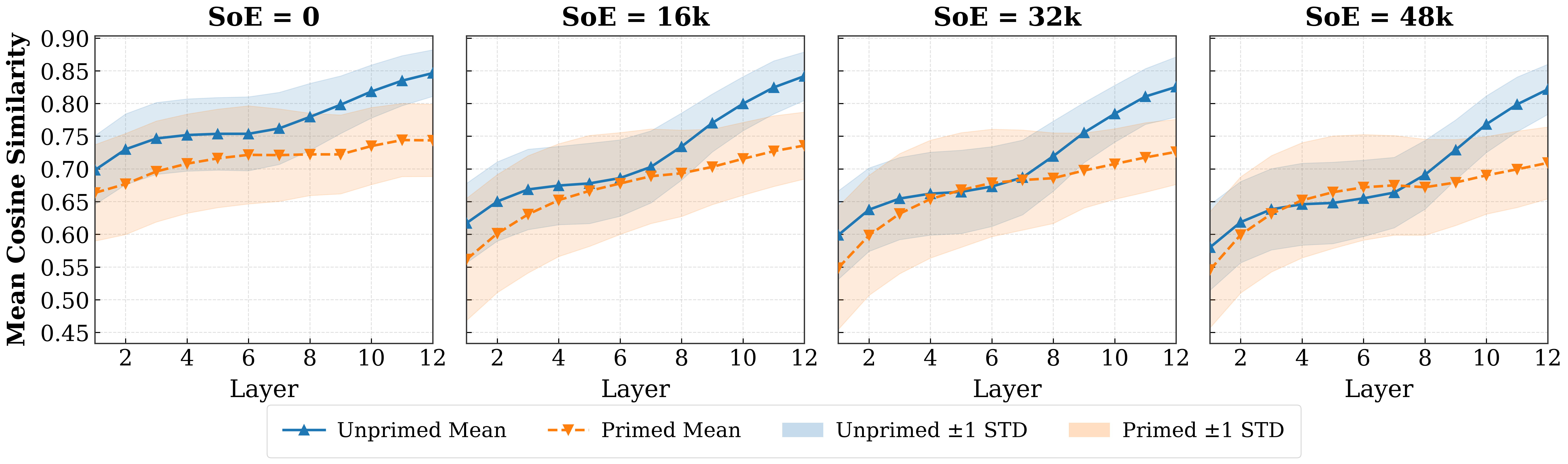}
    \caption{Layer-wise raw mean cosine similarity of surprisal differences across all language pairs for the four Step of Exposure (SoE) checkpoints. The shaded regions denote $\pm 1$ standard deviation. In the unprimed condition, representations converge tightly in the later layers. In contrast, explicit priming forces the network to maintain distinct, structurally localized representations, an effect that strengthens (lowering overall similarity) as L1 dominance increases.}
    \label{fig:cosine_sim}
\end{figure*}

Figure~\ref{fig:cosine_sim} confirms our hypothesis regarding the suppression of L1-specific features and the subsequent convergence toward a universal L2 representation in the unprimed condition. The raw mean cosine similarity across language pairs increases sharply starting around layer 6, ultimately reaching up to 0.85 in the terminal layers. This structural convergence precisely matches the point where the correlation between similarity and linguistic distance shifts towards becoming insignificant in Figure~\ref{fig:surprisal_diff_corr}. Furthermore, increasing the Step of Exposure (SoE) lowers the overall cosine similarity in the unprimed condition, reflecting deeper latent entrenchment of the respective L1s. 

Explicit priming drives this similarity even lower and fundamentally alters the layer-wise trajectory. Unlike the unprimed baseline, priming forces the deeper layers to maintain distinct, L1-dependent representations. In the primed condition, the maximum cosine similarity plateaus at approximately 0.75 for SoE=0 and steadily decreases at higher SoE checkpoints. This demonstrates that increased L1 dominance permanently anchors the deeper-layer representations to the specific structural properties of the L1 prime, preventing the network from converging back to a clean monolingual L2 state.

\subsection{Neuron Analysis}
\label{sec:neuron_analysis}
\subsubsection{Linguistic Distance and Shared Neural Activations for L2 Syntax}
\label{sec:neuron_distance}
To provide mechanistic evidence for crosslinguistic transfer, we analyze the grammar neurons identified in Section~\ref{sec:finding_neurons}. While previous research has investigated cross-lingual neuron sharing within  multilingual models \cite{varda-marelli-2023-data, kryvosheieva2026differenttypessyntacticagreement, mueller-etal-2022-causal}, our study isolates this phenomenon by analyzing independent bilingual models trained across diverse L1 backgrounds. We specifically investigate whether structurally similar L1s cause the network to route L2 grammatical information through shared neural pathways. 

Figure~\ref{fig:neurons_vs_distance} plots the number of shared active neurons against linguistic distance for all possible L1 pairs ($N=105$). To account for random neuron overlap due to initialization or other artifacts of this analysis, we apply our neuron identification methodology to untrained models initialized with identical random seeds. We then use this random overlap as a control baseline to compute a partial Spearman correlation. To further contextualize these results, we calculated the partial correlation of neuron overlap and linguistic distance in independent monolingual models. We find that this monolingual baseline correlation is statistically insignificant in both the unprimed ($\rho_{\text{part}}=-0.01$) and primed ($\rho_{\text{part}}=-0.15$) conditions ($p>0.05$ in both cases). Overall, we observe that languages share significantly fewer neurons in the unprimed condition compared to the primed condition. Furthermore, the correlation between shared neural overlap and syntactic distance is heavily dependent on the Step of Exposure (SoE). In the balanced bilingual setting (SoE=0), the correlation in the unprimed condition is near zero and not statistically significant. However, as the SoE increases, this correlation rapidly strengthens and becomes significant. This progression demonstrates that early L1 entrenchment acts as a necessary catalyst for structurally consistent crosslinguistic influence at the unit level.

Beyond these correlational shifts, Figure~\ref{fig:neurons_vs_distance} reveals important architectural dynamics regarding the size of the shared bilingual network. First, a persistent volume gap exists between the two conditions across all training stages. The primed condition consistently recruits a significantly higher absolute number of shared neurons than the unprimed baseline. Second, as the SoE increases, the absolute number of shared neurons systematically drops across both conditions. This can be explained by L2 language proficiency: as the SoE increases, the number of steps for training on the L2 data shrinks, which can lead to sparser neuron activation.

\begin{figure*}[t!]
    \centering
    \includegraphics[width=1.\textwidth]{figures/neurons/language_neurons_vs_distance.png}
    \caption{The relationship between syntactic distance and the number of shared active neurons across 105 language pairs. The analysis is presented for both unprimed and primed conditions across four Step of Exposure (SoE) checkpoints. The progression of the partial Spearman correlation ($\rho\_{\text{part}}$) demonstrates that increased L1 dominance (higher SoE) strengthens the typological alignment of shared neural representations.}
    \label{fig:neurons_vs_distance}
\end{figure*}

While both conditions develop a significant negative correlation with syntactic distance over time, their trajectories diverge. The unprimed correlation stabilizes relatively quickly, hovering between $\rho=-0.23$ and $\rho=-0.26$ in the later steps. In contrast, the primed correlation steepens continuously, reaching $\rho=-0.37$ at SoE=48k. This divergence indicates that L1 dominance disproportionately sensitizes the explicit priming mechanism to typological distance, whereas the latent interference in the unprimed condition hits a ceiling.

Crucially, these unit-level results provide mechanistic support for our earlier behavioral findings. The lack of a significant correlation between syntactic distance and shared neurons at SoE=0 perfectly mirrors the absence of correlation observed in our surprisal difference analysis (Figure~\ref{fig:surprisal_diff_corr}). By demonstrating that the priming effect is structurally embedded in the neuronal allocation of the models rather than manifesting as arbitrary interference, this analysis confirms that L1 dominance fundamentally modulates both the magnitude and the typological coherence of crosslinguistic influence.

\subsubsection{The Per-Layer Number of Neurons in Primed and Unprimed Models}
\label{sec:neuron_layer}
To further investigate the internal structural mechanics of this transfer, we examine the physical distribution of identified grammar units across the model layers. Rather than looking strictly at representational overlap between language pairs, we calculate the total absolute number of grammar neurons isolated within each individual transformer layer for both the primed and unprimed conditions. 

Figure~\ref{fig:layerwise_neurons} displays these layer-wise distributions across all 15 bilingual configurations alongside the collective averages. In the unprimed condition, grammar units follow a clear bell-shaped distribution that peaks in the middle layers, specifically between layers 5 and 7. This mid-level concentration aligns with established findings in the interpretability literature showing that core syntactic properties are primarily handled within the intermediate layers of a transformer \cite{jawahar-etal-2019-bert, tenney-etal-2019-bert}. 

Crucially, explicit priming induces a dramatic structural migration, shifting the peak concentration of grammar units into the deepest network layers (predominantly layers 10 and 11). Rather than merely modulating existing intermediate representations, priming systematically defers the resolution of grammatical dependencies to these upper layers, where long-range context is consolidated. This topographical migration coincides precisely with the emergence of distinct layer-wise priming effects (Figure~\ref{fig:layer_prime_effect}) and the point at which the correlation between surprisal and linguistic distance becomes statistically significant (Figure~\ref{fig:surprisal_diff_corr}). Together, these results demonstrate that late-layer crosslinguistic influence perfectly aligns with the physical localization of active grammatical units, confirming that explicit priming induces a structured, localized computational shift rather than introducing arbitrary network noise. In Section \ref{sec:intervention_components} we ablate the effect on intervening on the identified neurons, finding that deactivating syntactic neurons hurts grammatical accuracy significantly more than deactivating a randomly drawn equal number of neurons (i.e. up to 3\% compared to <1\% respectively).

\begin{figure*}[t!]
    \centering
    \includegraphics[width=1.\textwidth]{figures/neurons/layerwise_number.png}
    \caption{Layer-wise distribution of identified grammar units across all 15 bilingual model variations, evaluated under both unprimed (solid lines) and primed (dashed lines) conditions. The bold black trendlines denote the cross-lingual averages across the four Step of Exposure (SoE) checkpoints. The results demonstrate a clear spatial migration, where explicit priming shifts the primary hub of grammatical processing from the middle layers to the upper terminal blocks of the network.}
    \label{fig:layerwise_neurons}
\end{figure*}

\subsubsection{The Influence of Typological Distance on Primed and Unprimed Neural Overlap}
\label{sec:distance_overlap}
In psycholinguistic theory, structural priming is often explained by the activation of shared or densely connected representations across languages \cite{hartsuiker2004, HARTSUIKER_BERNOLET_2017, vasilyeva2010priming, danbi2024separate}. A crucial question is whether an analogous mechanism exists within artificial language models. As demonstrated in Figure~\ref{fig:layerwise_neurons}, explicit priming induces a topological shift in neuron density toward later layers. Consequently, the absolute number of overlapping neurons between the primed and unprimed states is a relatively small fraction of the total identified syntactic pool (i.e., 3,223 units\footnote{Calculated as 5\% of the query, value, and output attention neurons with 768 units each, plus 5\% of the 3,072 neurons in the MLP block in a 12 layer transformer.}). The layer-wise migration shifts the primary locus of grammatical processing to the terminal layers, naturally limiting intersection with the mid-layer unprimed representations. However, a functionally significant portion of active neurons remains overlapping. In this section, we investigate this specific overlap to determine whether L1s that are typologically similar to the L2 activate a higher number of shared neurons with the unprimed L2 baseline than distant L1s.

\begin{figure*}[t!]
    \centering
    \includegraphics[width=1.\textwidth]{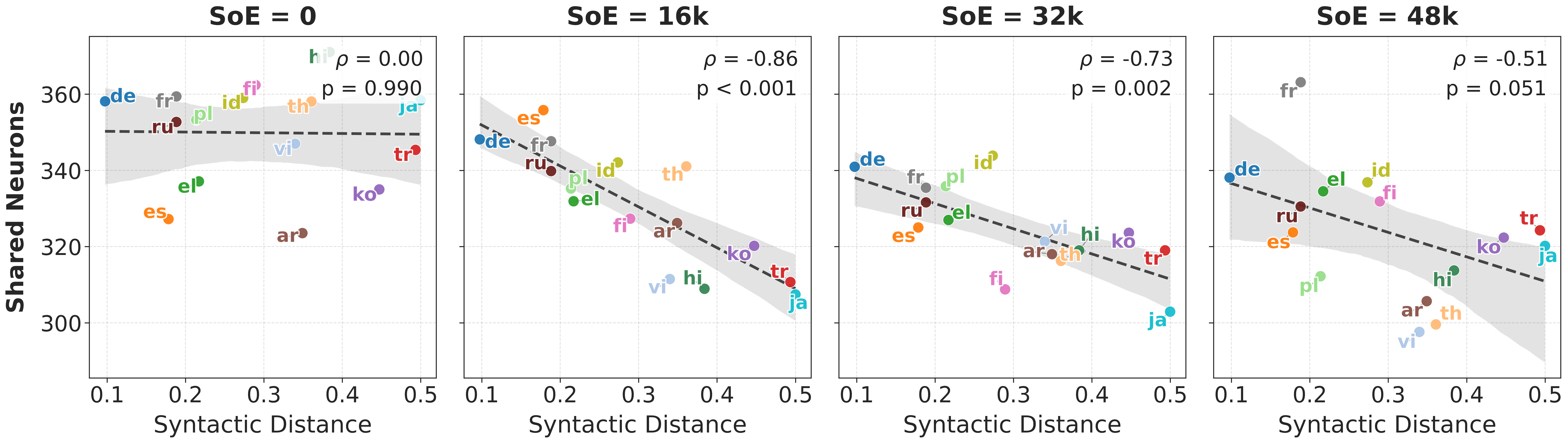} 
    \caption{Scatter plots showing the number of overlapping active neurons between the primed and unprimed conditions for each of the 15 bilingual models ($N=15$), plotted against the linguistic distance of the L1. The panels demonstrate how the Spearman correlation ($\rho$) evolves across the four Step of Exposure (SoE) checkpoints, highlighting a peak in structural alignment at SoE=16k followed by a gradual weakening as L2 proficiency declines in the later stages.}
    \label{fig:distance_vs_overlap}
\end{figure*}

Figure~\ref{fig:distance_vs_overlap} illustrates the number of shared neurons between the primed and unprimed states for each L1 model as a function of linguistic distance. Consistent with our earlier findings (Figures~\ref{fig:surprisal_diff_corr} and~\ref{fig:neurons_vs_distance}), the correlation at SoE=0 is not statistically significant. In this balanced bilingual setting, the equivalence of language dominance leads to a high volume of shared neural activation regardless of the specific L1 prime. However, as L1 dominance increases at SoE=16k, a strong negative correlation emerges ($\rho = -0.86$). This indicates that under moderate L1 dominance, typological proximity directly dictates the extent to which the prime activates the established L2 neural circuitry.

Surprisingly, as SoE increases further, this correlation weakens, dropping to $\rho = -0.51$ at SoE=48k. This contrasts with the crosslinguistic consistency between different L1 models (Figure~\ref{fig:neurons_vs_distance}), which strictly strengthens with higher SoE. This divergence suggests a mechanistic trade-off. While higher L1 dominance makes the explicit priming mechanism more typologically coherent across different L1s, the accompanying reduction in L2 exposure (due to fewer L2 training steps at higher SoEs) degrades the unprimed L2 representations. Consequently, the L2 grammatical neurons become sparser and less robust, making them harder to consistently activate even by a structurally similar L1 prime. This interpretation is further supported by the steady decline in the absolute number of shared neurons at higher SoE checkpoints. 

As shown in Figure \ref{fig:prime_effect}, overall BLiMP performance in the primed models is strongly negatively correlated with linguistic distance, exhibiting a decrease in positive transfer and a sharp increase in negative transfer as structural divergence grows. This raises the critical question of whether the successful activation of overlapping neurons is functionally necessary for preserving downstream accuracy. Intuitively, recruiting these shared units might not elevate performance beyond the unprimed L2 baseline, but it likely mitigates severe degradation by maintaining the unprimed model structures. To isolate this causal effect and verify whether the activation of these specific intersecting neurons explains the mitigation of interference, we conduct a targeted ablation study in Section \ref{sec:intevention_overlap}.

\section{Ablation Studies}

\subsection{Evaluating Bilingual Training Configurations}
\label{sec:ablation_training}

Our core bilingual training protocol, outlined in Section~\ref{subsec:bilingual_training}, relies on specific architectural and curriculum choices. In this section, we systematically ablate four of these training factors to evaluate their precise impact on crosslinguistic influence. We quantify the priming effect by measuring the ratio of predictions that change relative to the unprimed baseline. Positive transfer is defined as a shift from an incorrect to a correct prediction, whereas negative transfer is a shift from correct to incorrect. We calculate Wilcoxon signed-rank test to determine if the distribution of delta deviates significantly from 0. We select this as alternative to the standard paired t-test because our sample size of evaluation languages is relatively small ($N=15$). To manage computational costs, we report the cross-lingual averages ($N=15$) exclusively for the SoE=16k and SoE=32k checkpoints. 

\begin{figure*}[t!]
    \centering
    \includegraphics[width=1.\textwidth]{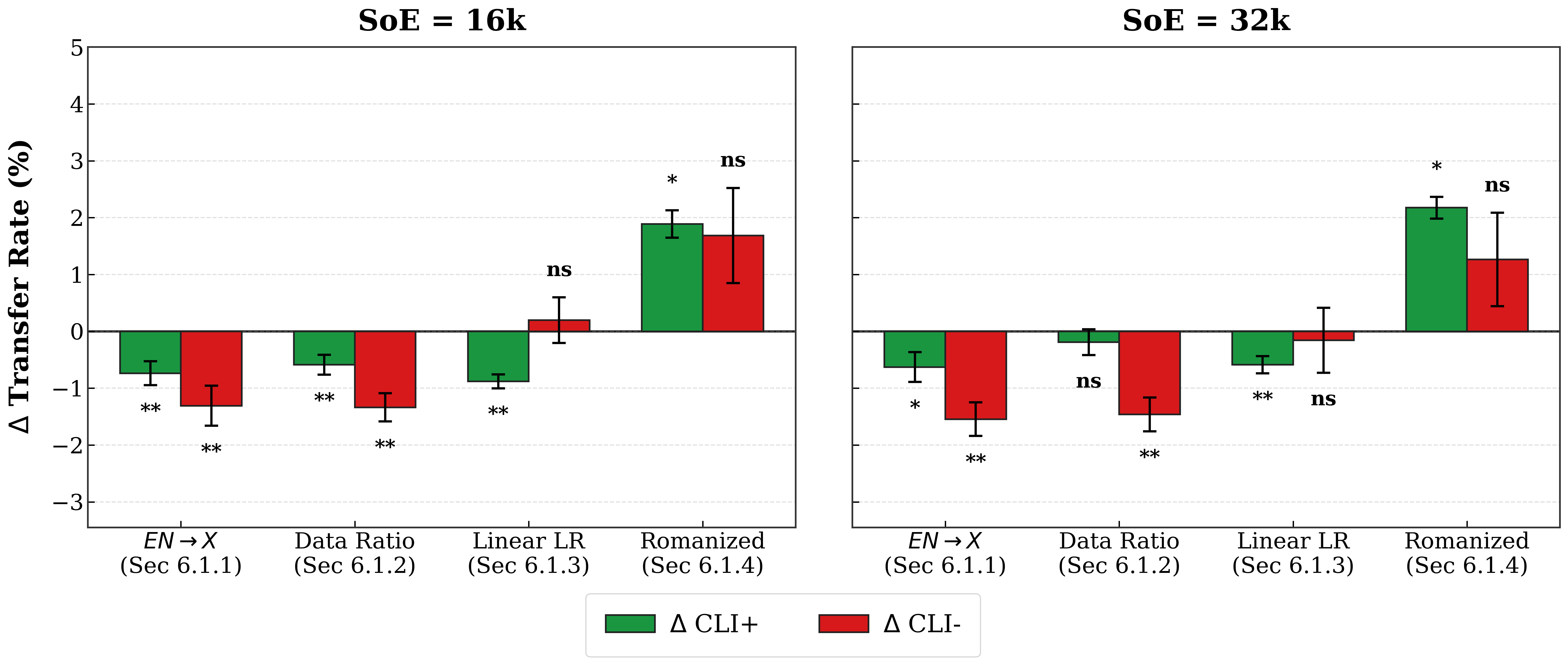}
    \caption{Difference in transfer rates ($\Delta \text{CLI}^+$  and $\Delta \text{CLI}^-$) across the four ablation conditions relative to the standard bilingual baseline models. Results are averaged across the evaluated language pairs at the SoE=16k and SoE=32k checkpoints. Positive values indicate an increase in the respective transfer rate compared to the baseline. Statistical significance is denoted by ** ($p < 0.01$) and * ($p < 0.05$), while 'ns' indicates not significant ($p > 0.05$).}
    \label{fig:ablation_training}
\end{figure*}

We investigate the bidirectionality of priming (Section~\ref{sec:asymetry}), the impact of sequential versus simultaneous data exposure (Section~\ref{sec:ablate_data}), the influence of learning rate schedulers (Section~\ref{sec:ablate_linear}), and the effect of script Romanization (Section~\ref{sec:romanization}). The aggregate results are presented in Figure~\ref{fig:ablation_training} as the delta transfer rates relative to our primary experimental models.

\subsubsection{Bidirectionality of Priming Effects}
\label{sec:asymetry}

To determine whether crosslinguistic influence operates symmetrically, we train reverse-curriculum models where English serves as the L1 and the 15 other languages function as the respective L2s (En-X). We then evaluate priming in the reverse direction by using the L2 languages to prime the L1 which now is English. Figure~\ref{fig:ablation_training} displays the delta positive and negative priming rates relative to the standard models (X-En). The results indicate that when English is the dominant L1 and the priming language is the L2, the models exhibit lower positive and negative transfer rates compared to when English is the L2. This asymmetry suggests that the direction of language dominance fundamentally modulates crosslinguistic influence. Specifically, the more dominant language leads to stronger priming effects. 

\subsubsection{Simultaneous vs. Sequential Data Exposure}
\label{sec:ablate_data}

Our primary curriculum trains the L1 sequentially before the L2, a choice hypothesized to establish early L1 entrenchment. To isolate the effect of this ordering, we train ablation models on the exact same unbalanced data distributions but expose the network to the L1 and L2 simultaneously. Because the learning rate remains constant, the primary variable is solely the order of exposure (except the warmup stage). Figure~\ref{fig:ablation_training} shows that simultaneous training yields a measurable drop in both positive and negative transfer compared to the sequential baseline. Notably, negative transfer decreases more sharply. This confirms our hypothesis that early, sequential exposure is a critical driver of L1 dominance. 

\subsubsection{Influence of the Learning Rate Scheduler}
\label{sec:ablate_linear}

In Section~\ref{subsec:bilingual_training}, we motivated the use of a constant learning rate with warmup to eliminate the confounding variable of a decaying learning rate when varying the SoE. To ensure this choice does not artificially degrade baseline language modeling capabilities, we compare our constant scheduler against a standard linear decay scheduler. Table~\ref{tab:model_comparison} reports the mean BLiMP accuracy for five monolingual English models trained using distinct bilingual tokenizers. The performance difference is negligible, demonstrating that the constant learning rate does not undermine general syntactic capability.

\begin{table}[htbp]
\centering
\begin{tabular}{lc}
\hline
\textbf{Scheduler} & \textbf{Accuracy (\%)} \\ \hline
Linear Scheduler       & $77.14 \pm 0.32$       \\
Constant Scheduler     & $77.08 \pm 0.35$       \\ \hline
\end{tabular}
\caption{Mean BLiMP accuracy comparing linear and constant learning rate schedulers for monolingual English models. Results are averaged across five models trained with different L1-English tokenizers (L1s: German, Spanish, Greek, Turkish, Korean).}
\label{tab:model_comparison}
\end{table}

Figure~\ref{fig:ablation_training} illustrates the transfer dynamics when models are instead trained with a linear scheduler. The linear scheduler yields comparable average negative transfer but noticeably lower positive transfer. Under a linear decay, the L2 is introduced during a phase of lower learning rates, which inhibits L2 entrenchment and proficiency. This relative structural weakness suppresses positive transfer, while negative transfer from the strongly entrenched L1 remains persistent.

\subsubsection{Impact of Script Romanization}
\label{sec:romanization}

Previous research indicates that unifying language scripts can facilitate cross-lingual transfer \cite{dhamecha-etal-2021-role, purkayastha-etal-2023-romanization, moosa-etal-2023-transliteration}. To investigate whether distinct scripts bottleneck structural priming, we romanize the datasets for the seven non-Latin languages in our study (Arabic, Greek, Hindi, Japanese, Korean, Russian, and Thai) using the Uroman tool \cite{hermjakob-etal-2018-box}. These languages span a wide range of linguistic distances from English (ranging from 0.19 for Russian to 0.50 for Japanese). We train new models from scratch using the romanized data and subsequently prime them with romanized L1 text and evaluate transfer effects to the L2. We compare these models to their equivalent that use the original script in Figure~\ref{fig:ablation_training}. Script divergence indeed acts as a representational bottleneck. Unifying the scripts significantly amplifies overall crosslinguistic influence, increasing both positive and negative transfer rates, with positive transfer experiencing the more substantial increase. The gap between positive and negative transfer widens as SoE increases from 16k to 32k, showing that a shared script improves the consistency in activation of shared structures as L1 dominance increases. The effect on negative transfer seems to be insignificant (p > 0.05) but this could be attributed to the low sample size in this ablation ($N=7$).

\subsection{Priming Ablation}
\subsubsection{Ablation of Prime Structure and Content}
\label{sec:perturbations}

To isolate the properties of the prime driving crosslinguistic influence, we compare using direct translation primes against four baselines: (1) \textbf{Shuffle}, a prime from the same grammatical category but with disjoint vocabulary, testing whether syntactic structure alone induces transfer; (2) \textbf{Random}, an unrelated prime from a different grammatical category; (3) \textbf{Scramble}, the direct translation prime with randomly reordered words, preserving lexical overlap but removing syntactic structure; and (4) \textbf{Coin Flip}, a baseline representing a 50\% random guessing probability. Results are averaged across five random seeds for both the SoE=16k and SoE=32k checkpoints. We calculate Wilcoxon signed-rank test to determine if the distribution of deltas (averaged over the random seeds) deviates significantly from 0. 

\begin{figure*}[t!]
\centering
\includegraphics[width=1.\textwidth]{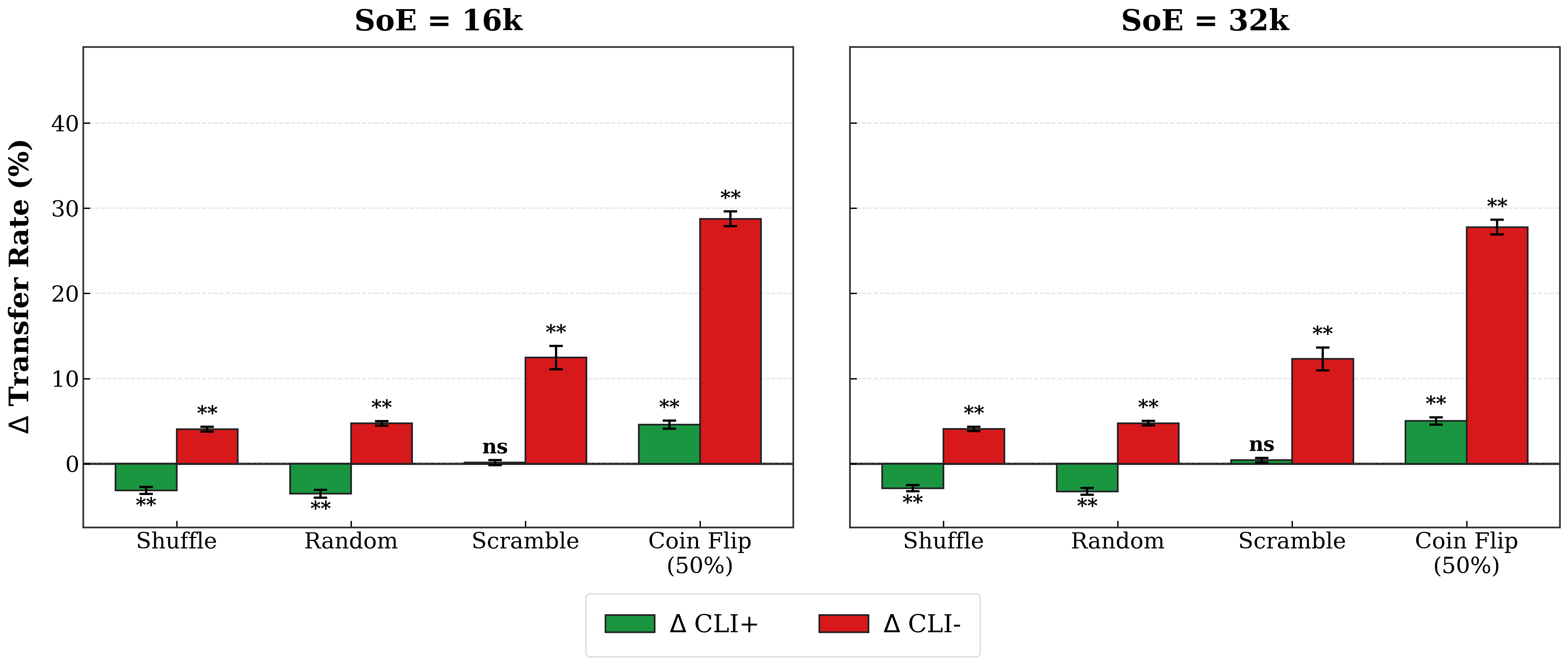}
\caption{Difference in transfer rates ($\Delta \text{CLI}^+$  and $\Delta \text{CLI}^-$) for four prime baselines relative to the standard direct translation. Results are averaged across 5 random seeds at the SoE=16k and SoE=32k checkpoints. Statistical significance is denoted by ** ($p < 0.01$) and * ($p < 0.05$), while 'ns' indicates not significant ($p > 0.05$).}
\label{fig:perturbations}
\end{figure*}

Figure~\ref{fig:perturbations} illustrates the shifts in positive and negative transfer rates relative to the direct translation baseline. Compared to direct translation, both the Shuffle and Random conditions yield reduced positive transfer and increased negative interference. The minimal performance variance between the Shuffle and Random conditions indicates that syntactic structure alone, devoid of lexical or semantic alignment, is insufficient to induce reliable transfer. Instead, direct translation provides the essential lexical and semantic grounding required to successfully facilitate structural mapping between L1 and L2 representations. This aligns with previous findings demonstrating that structural priming in language models is significantly amplified by semantic similarity and lexical overlap \cite{sinclair-etal-2022-structural}.

The Scramble baseline isolates the contribution of syntactic structure. While positive transfer remains stable under this condition, negative transfer increases by a factor of two compared to the Shuffle and Random conditions. This indicates that lexical overlap with absent structural coherence degrades model accuracy rather than inducing consistent priming.

The Coin Flip baseline contextualizes these results. Given the unprimed model's high L2 competence (accuracy $>70\%$), random guessing inherently degrades correct predictions, resulting in elevated negative transfer ($\sim +28\%$). Conversely, random guessing inadvertently corrects a subset of previously incorrect predictions, yielding an artificial increase in positive transfer ($\sim +5\%$). The behavioral profile of the Scramble condition reflects this dynamic. Providing disorganized tokens appears to confuse the model, resulting in predictions that resemble random guessing rather than systematic crosslinguistic transfer.

\subsubsection{Isolating the Effect of Prime Translation Quality}
\label{sec:translation_ablation}

Because we rely on automatic translation to generate the L1 primes, it is necessary to eliminate the confounding hypothesis that our crosslinguistic transfer results are merely artifacts of translation quality. Under this null hypothesis, primes would induce positive transfer simply when their translation accuracy is high and negative transfer when the translation is poor. While we restricted our study to high-resource languages within the NLLB model \cite{nllbteam2022languageleftbehindscaling}, cross-lingual disparities in translation quality naturally persist. 

To evaluate this, we measure the referenceless translation quality of the primes using COMET\footnote{Unbabel/wmt22-cometkiwi-da}. Because baseline translation scores vary intrinsically between languages, we compute the z-score of the COMET scores within each individual language and segment the data into five relative quality bins (1=Worst, 5=Best). We then measure the grammatical preference by calculating the difference in log-probability ($\Delta$ LogProb) between the acceptable and unacceptable BLiMP sentences across these bins. To isolate the effect of the prime, we compare the primed bilingual models against a monolingual L2 baseline that was never exposed to L1 training data or translation primes. 

\begin{figure*}[t!]
    \centering
    \includegraphics[width=1.\textwidth]{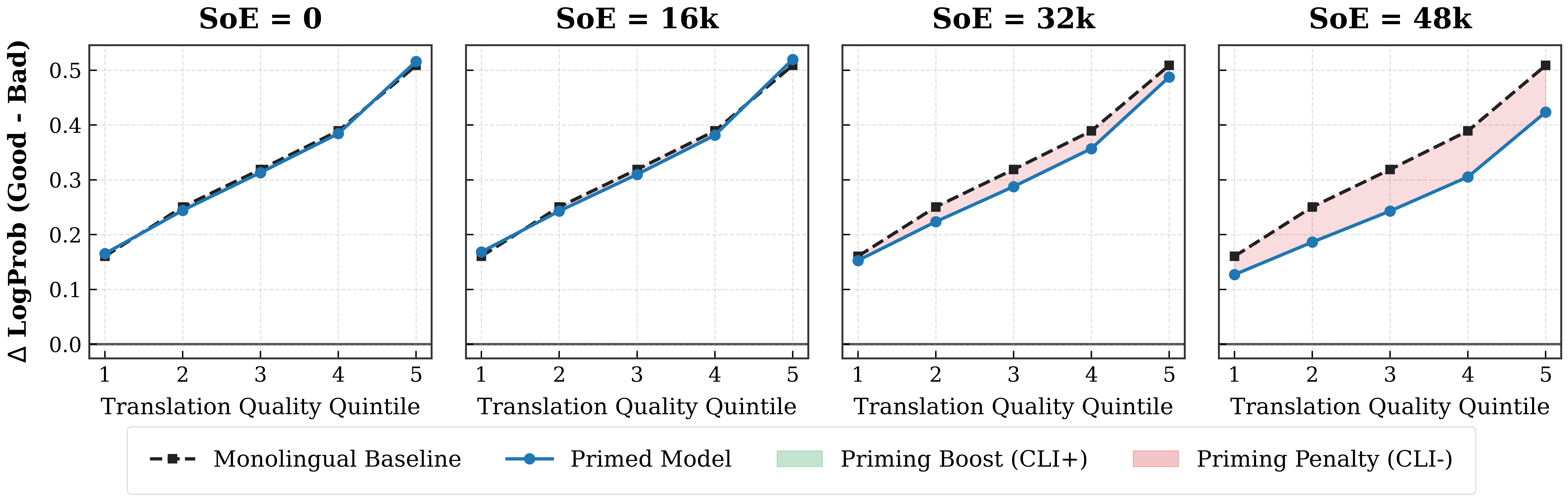}
    \caption{The effect of prime translation quality on the $\Delta$ LogProb across four Step of Exposure (SoE) checkpoints. Translation quality is measured via COMET and grouped into five relative bins (1=Worst, 5=Best). The shaded red region indicates a ``Priming Penalty'' where the primed bilingual model underperforms the monolingual L2 baseline. The widening of this penalty at higher translation qualities demonstrates that highly idiomatic L1 primes induce greater structural interference, refuting the hypothesis that transfer effects are merely artifacts of translation quality.}
    \label{fig:translation_qe}
\end{figure*}

Figure~\ref{fig:translation_qe} illustrates the average relationship between relative translation quality and $\Delta$ LogProb across all 15 languages. If poor translation quality were the primary driver of negative transfer, we would expect the performance gap between the primed model and the monolingual baseline to be largest at Bin 1 and to shrink or reverse into a positive boost at Bin 5. 

The data reveals the exact opposite trend. Initially, both the primed and monolingual models exhibit a strictly positive correlation between translation quality and grammatical preference. This shared upward trajectory indicates that sentences which are easier to translate are simply inherently easier for language models to parse grammatically. However, as the Step of Exposure (SoE) increases to 32k and 48k steps, a distinct ``Priming Penalty'' emerges, where the primed model underperforms the monolingual baseline. Strikingly, this penalty widens precisely as translation quality increases. 

This widening gap offers evidence against the translation-quality confound. State-of-the-art neural metrics like COMET are generally designed to penalize ``translationese'' and reward highly fluent, idiomatic text. Consequently, it is possible that the highest quality L1 primes are those that most rigidly enforce L1-specific syntactic structures. As established in our earlier analyses of L1 entrenchment, higher SoE appears to increase the model's sensitivity to these structures. Therefore, priming a highly dominant L1 model with an idiomatic L1 sentence maximize the structural interference imposed on the L2 target, potentially leading to a larger drop in accuracy. This pattern suggests that underlying typological compatibility, rather than superficial translation quality, plays a primary role in crosslinguistic influence. We present further support for these observations in Appendix~\ref{sec:translation_appendix}, demonstrating that this trend holds across individual languages and replicates using an alternative referenceless metric (MetricX-XL\footnote{google/metricx-24-hybrid-xl-v2p6}).

\subsubsection{Target Language Susceptibility and Reverse Priming}
\label{sec:multiblimp}
In Section~\ref{sec:asymetry}, we demonstrated the bidirectionality of crosslinguistic influence by alternating English as the L1 and L2. However, because English remained the evaluation target in both configurations, it is possible that the observed symmetry is an artifact of English being uniquely susceptible to structural priming. For instance, \citet{arnett-etal-2025-acquisition} observed stronger structural priming effects when English served as the target rather than the source, suggesting that asymmetric transfer might be an intrinsic property of the target language itself. 

To isolate the effect of the target language, we conduct a reverse priming evaluation where English serves as the prime and the other languages serve as the targets. We utilize the MultiBLiMP dataset \cite{jumelet-etal-2026-multiblimp}, a multilingual minimal pair benchmark. From its 101 supported languages, we evaluate the 10 overlapping languages present in our study (Arabic, Finnish, French, German, Greek, Hindi, Polish, Russian, Spanish, and Turkish). To ensure a precise comparison with the English BLiMP baseline, we restrict the evaluation to subject-verb agreement phenomena (marked SV-\#) and exclude subject-participle agreement pairs. We then translate the acceptable target sentences into English using NLLB-3.3B to serve as the primes. For a fair comparison, the English BLiMP baseline in this ablation is similarly filtered to only include subject-verb agreement categories and is averaged exclusively over the models trained with these exact 10 counterpart languages.

\begin{figure*}[t!]
    \includegraphics[width=1.\textwidth]{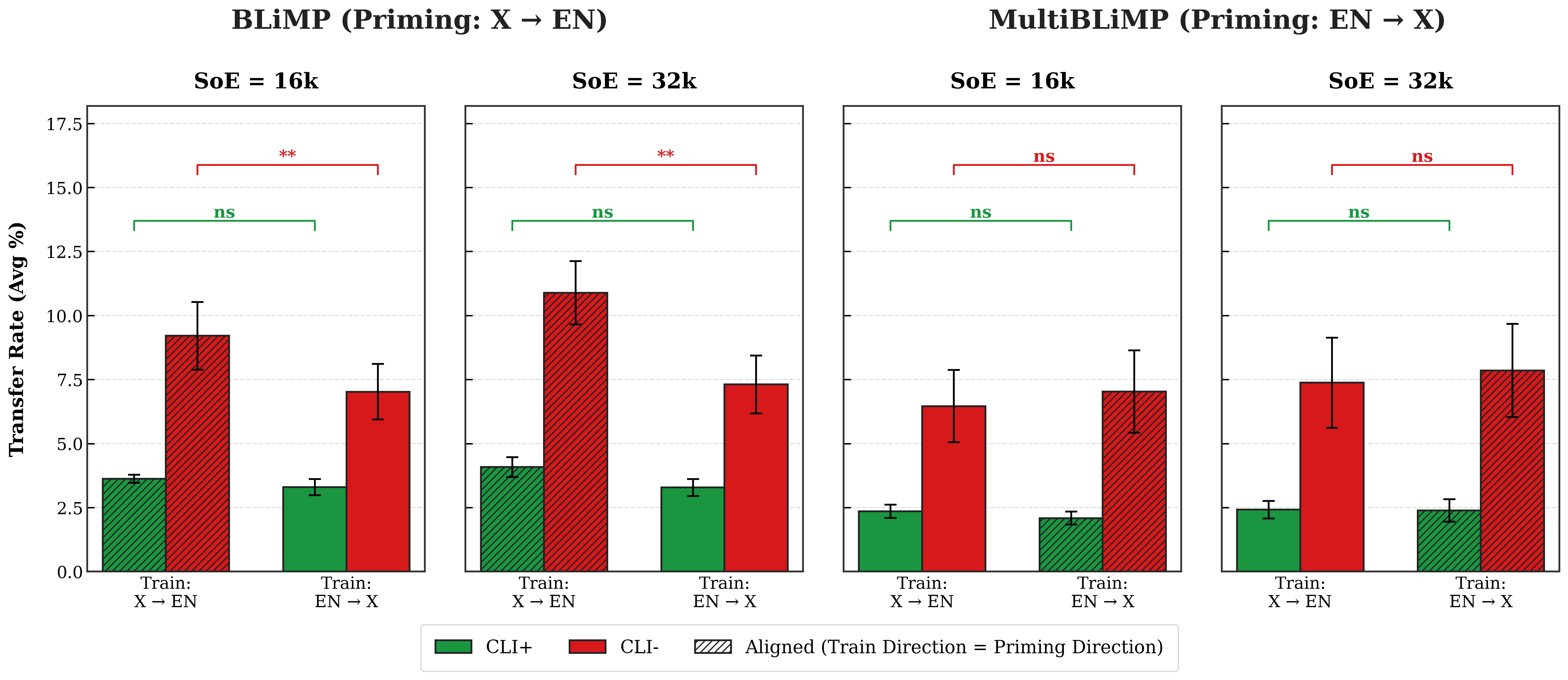}
    \caption{Comparison of positive and negative transfer rates when English is the target language (top row) versus when English is the priming language (bottom row). Hatched bars denote aligned configurations where the dominant L1 primes the L2. Statistical significance between aligned and unaligned configurations is determined via Wilcoxon signed-rank test (** indicates $p < 0.01$, ns indicates not significant). While crosslinguistic influence is strictly bidirectional, English exhibits a greater overall susceptibility to both positive and negative transfer when acting as the target. Furthermore, L1 dominance significantly exacerbates negative transfer only when English is the target language, while having no significant effect on positive transfer in either direction.}
    \label{fig:multiblimp}
\end{figure*}

Figure~\ref{fig:multiblimp} contrasts the transfer rates when English is the target language against when the other 10 languages are the targets. The results confirm that structural priming is fundamentally bidirectional regardless of the target. However, the overall magnitude of both positive and negative transfer is noticeably stronger when English is the target language. 

Furthermore, we examine the role of language dominance in this reverse setting. The hatched bars in Figure~\ref{fig:multiblimp} indicate aligned configurations where the training direction matches the priming direction (i.e., the dominant L1 primes the L2). To evaluate this rigorously, we conduct Wilcoxon signed-rank test comparing the transfer rates of the aligned configurations against their unaligned counterparts. The analysis reveals that priming from the dominant L1 significantly amplifies negative transfer ($p < 0.01$) only when English is the target language. When non-English languages serve as the target, the slight increase in negative transfer is not statistically significant. Regarding positive transfer, aligning the prime with the dominant L1 yields no statistically significant improvement in either direction.

Ultimately, these findings corroborate the observations of \citet{arnett-etal-2025-acquisition}, confirming that structural priming operates with greater magnitude when English serves as the target. Furthermore, the capacity for L1 dominance to significantly amplify negative transfer appears uniquely sensitive to English targets. While we cannot entirely rule out the possibility that some of this asymmetry stems from inherent distributional and methodological differences between the BLiMP and MultiBLiMP benchmarks, the primary conclusion stands: whether English is the target or the source of priming, the priming effect remains robust and bidirectional.

\subsection{Neuron Intervention}
\label{sec:neuron_intervetion}

We perform a causal intervention to study whether neurons identified and analyzed previously are actually causally important for downstream accuracy. We follow previous work \cite{alkhamissi-etal-2025-llm, kryvosheieva2026differenttypessyntacticagreement} and achieve this by zeroing out the identified neurons and compare to ablating a control group of random neurons sampled with 5 different seeds. We then compare the BLiMP accuracy between ablating these two. The results are averaged across all BLiMP categories and L1 languages. In Section \ref{sec:intervention_components} we compare between ablating different neuron locations and in \ref{sec:intevention_overlap} we ablate the overlapping neurons between primed and unprimed models. 

\subsubsection{Targeted Component Ablation and the Role of Attention}
\label{sec:intervention_components}

\begin{figure*}[t!]
    \centering
    \includegraphics[width=1.\textwidth]{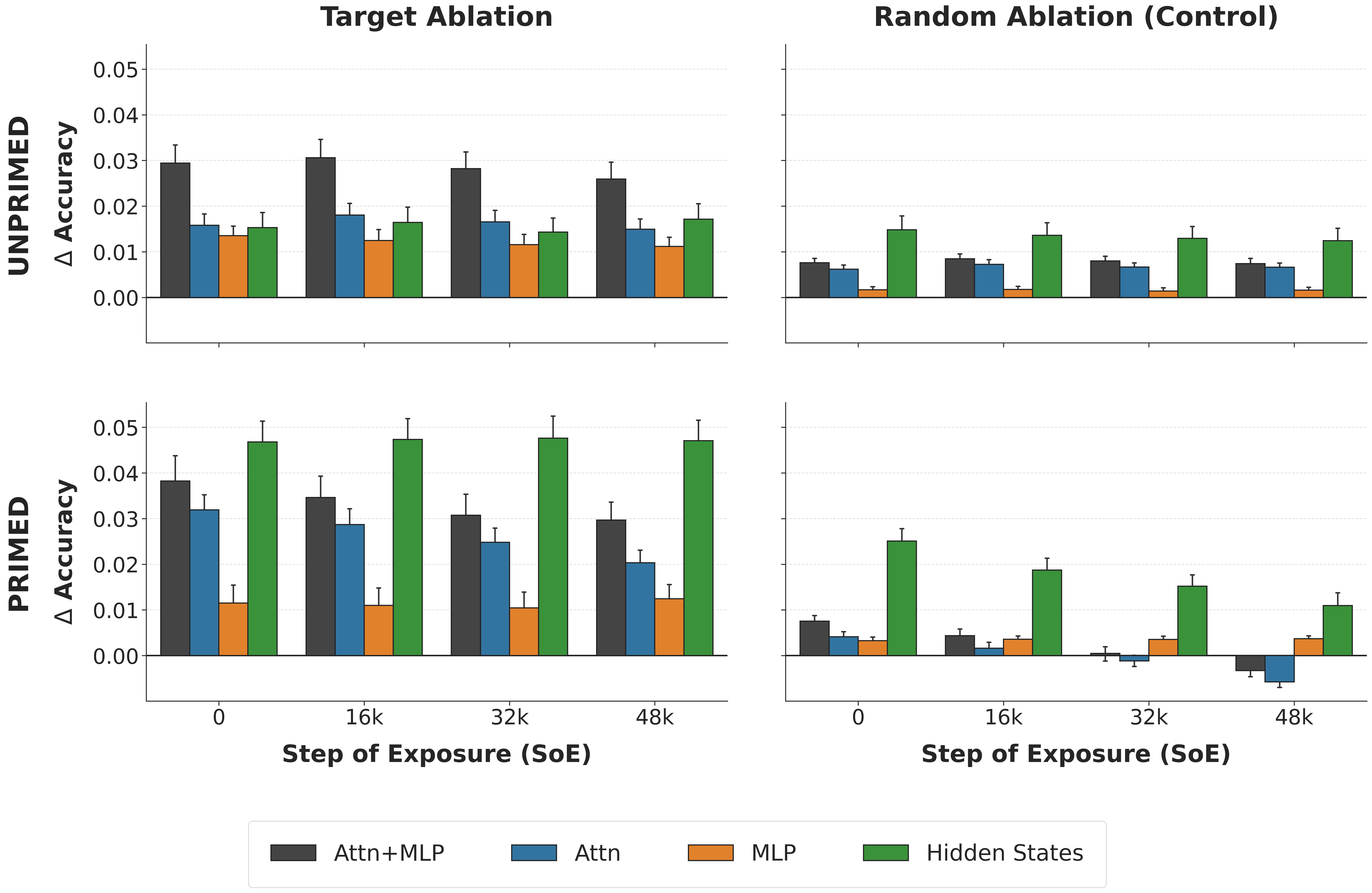}
    \caption{The effect of targeted neural ablations on BLiMP accuracy. The right panels show the accuracy drop when ablating identified syntactic units across different components (Attn+MLP, Attn only, MLP only, and Hidden States). The left panels show the corresponding control baseline where random units from those same components are ablated. Results are separated into unprimed (top) and primed (bottom) conditions across four Step of Exposure (SoE) checkpoints.}
    \label{fig:neuron_ablation}
\end{figure*}

To determine which specific architectural components govern crosslinguistic influence, we systematically ablate identified grammar units across different network modules. Figure~\ref{fig:neuron_ablation} presents the impact of ablating targeted syntactic neurons (right panels) versus randomly sampled neurons (left panels) within the unprimed (top) and primed (bottom) conditions. We compare our primary method (\textbf{Attn+MLP}) against isolated ablations of Attention (\textbf{Attn}) and Feed-Forward (\textbf{MLP}) neurons. Additionally, we include a baseline that ablates the macroscopic hidden states at the output of each transformer block, matching the intervention applied in previous work \cite{alkhamissi-etal-2025-llm, kryvosheieva2026differenttypessyntacticagreement}.

In the unprimed condition, ablating the combined \textbf{Attn+MLP} units yields the most substantial targeted accuracy drop ($\sim$3\%), with Attention neurons proving slightly more critical than MLP neurons. Crucially, while ablating the targeted macroscopic hidden states produces a drop in accuracy, this drop is virtually indistinguishable from the degradation caused by ablating random hidden states. In contrast, ablating random \textbf{Attn+MLP} units consistently results in an accuracy drop of less than 1\%. This indicates that macroscopic hidden state ablations in the unprimed model indiscriminately degrade general performance rather than isolating grammatical knowledge. Targeting granular neurons therefore provides a much more stable and precise intervention.

However, in the primed condition, the impact of ablating targeted hidden states surges to a nearly 5\% accuracy drop, outperforming the \textbf{Attn+MLP} intervention. We attribute this to the network's self-repair mechanism \cite{mcgrath2023hydraeffectemergentselfrepair}. While unprimed models leverage subsequent layers to recover from early-layer perturbations, explicit priming forces syntactic neurons into the terminal layers. This structural migration leaves no architectural depth for self-repair, rendering macroscopic ablations highly destructive. Consequently, this late-layer sensitivity also elevates the baseline impact of ablating random hidden states.

Furthermore, under explicit priming, the functional gap between targeted Attention and MLP neurons widens. While the importance of Attention units surges, the impact of MLP units remains relatively static. Interestingly, ablating random Attn and Attn+MLP units at higher SoE stages (32k and 48k) in the primed condition actually results in a negative accuracy drop (a negligible performance gain), underscoring just how highly specialized the targeted grammatical subset has become. Because attention mechanisms are fundamentally responsible for integrating long-range context, we hypothesize that they serve as the primary conduit for modulating crosslinguistic transfer from the L1 prime. If true, ablating these attention neurons in models with a typologically similar L1 should hurt performance by removing positive transfer, whereas ablating them in models with a distant L1 should mitigate interference.

We explicitly test this hypothesis in Figure~\ref{fig:delta_acc_distance} by visualizing the shift in ablation importance ($\Delta$ Accuracy Primed $-$ $\Delta$ Accuracy Unprimed) across linguistic distances. The results definitively isolate the source of the transfer. Starting at SoE=16k, a statistically significant negative correlation emerges exclusively for the Attention units. In stark contrast, the MLP units exhibit no significant correlation at any SoE stage. Consistent with our earlier findings on L1 entrenchment (Figure \ref{fig:neurons_vs_distance}), the correlation for Attention units strengthens progressively as the Step of Exposure increases, reaching $\rho = -0.73$ at SoE=48k. This provides direct causal evidence that the attention mechanism modulates crosslinguistic influence by attending to the L1 context, successfully leveraging the structural prior for similar languages while enforcing detrimental interference from distant ones.

\begin{figure*}[t!]
    \centering
    \includegraphics[width=1.\textwidth]{figures/neurons/delta_acc_distance.png}
    \caption{Scatter plots illustrating the shift in ablation importance ($\Delta$ Accuracy Primed $-$ $\Delta$ Accuracy Unprimed) for Attention versus MLP units relative to linguistic distance. A negative correlation indicates that ablating the component removes positive transfer for similar L1s while mitigating negative transfer for distant L1s. The significant correlations observed exclusively in the Attention units demonstrate that crosslinguistic influence is driven by attention mechanisms, an effect that strengthens with increased L1 entrenchment (higher SoE).}
    \label{fig:delta_acc_distance}
\end{figure*}

\subsubsection{Causal Evaluation of Overlapping Syntactic Neurons}
\label{sec:intevention_overlap}

\begin{figure*}[t!]
    \includegraphics[width=0.62\textwidth]{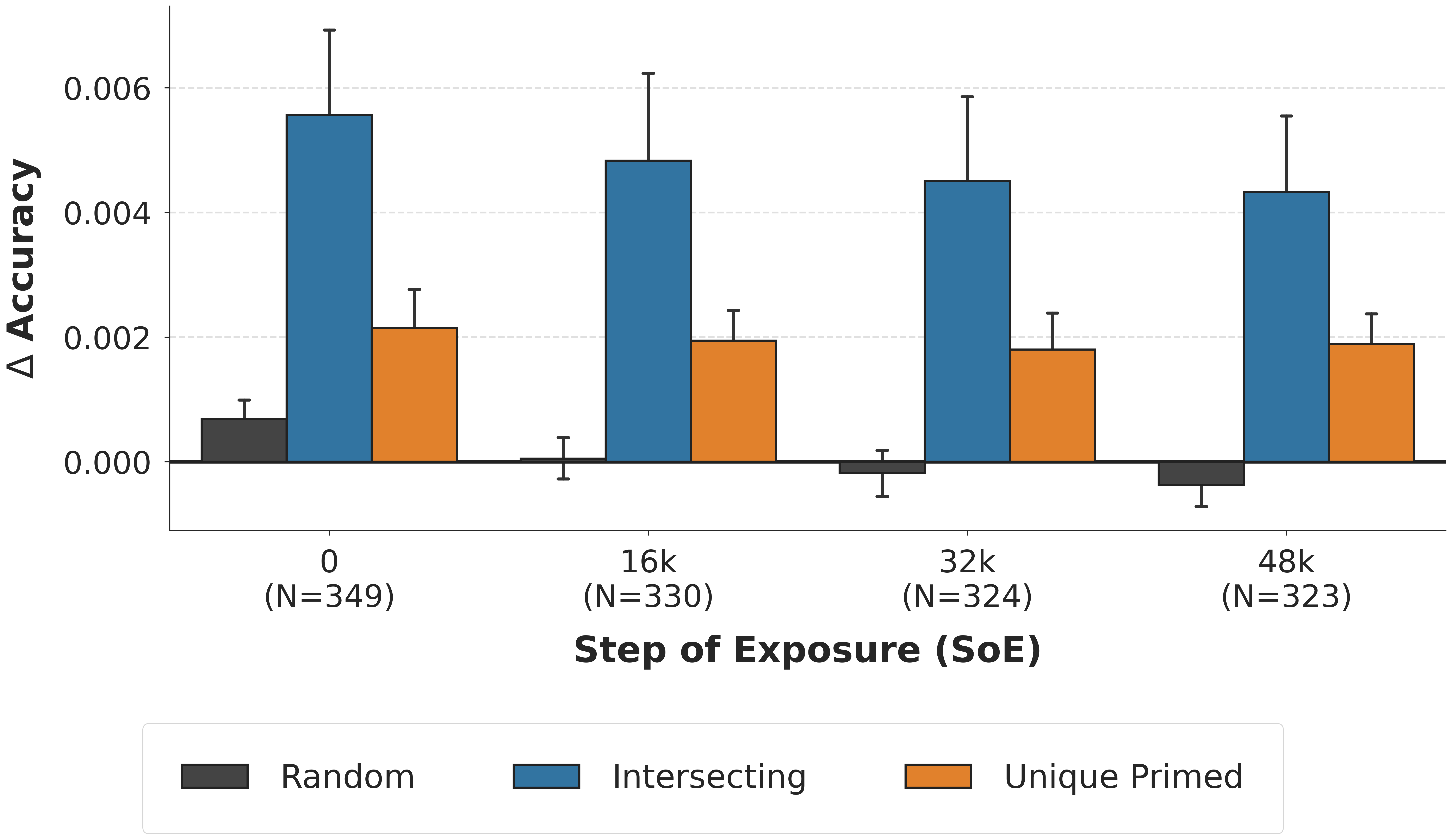}
    \caption{The effect of ablating distinct neuron subsets on BLiMP accuracy in the primed condition. Deactivating the intersecting neurons (shared between primed and unprimed states) causes a significantly larger accuracy drop than ablating unique prime-induced neurons or random baseline units. The total number of ablated intersecting neurons at each Step of Exposure (SoE) checkpoint is noted on the x-axis, highlighting a gradual reduction in the shared grammatical workspace over time.}
    \label{fig:ablate_overlap}
\end{figure*}

Our analysis in Section \ref{sec:distance_overlap} shows that the number of overlapping neurons between the primed and unprimed models is correlated with the linguitic distance between the L1 and L2, suggesting that priming with similar L1 languages preserves the unprimed model structures. A critical mechanistic question is whether these intersecting neurons (units active in both the primed and unprimed states) are functionally necessary for L2 grammatical resolution. To test this, we conduct a targeted ablation study. As shown in Figure~\ref{fig:ablate_overlap}, we compare the impact of deactivating these overlapping neurons against two control sets: an equivalent number of randomly sampled unique neurons (syntactic units active during priming but not in the unprimed state) and random non-syntactic neurons averaged across 5 seeds. 

The findings reveal that ablating the intersecting neurons induces more than double the accuracy drop compared to deactivating the unique prime-induced units across all four SoE conditions. This provides direct causal evidence that the capacity to route information through shared L2 neural structures mitigates interference.

\begin{figure*}[t!]
    \centering
    \includegraphics[width=1.\textwidth]{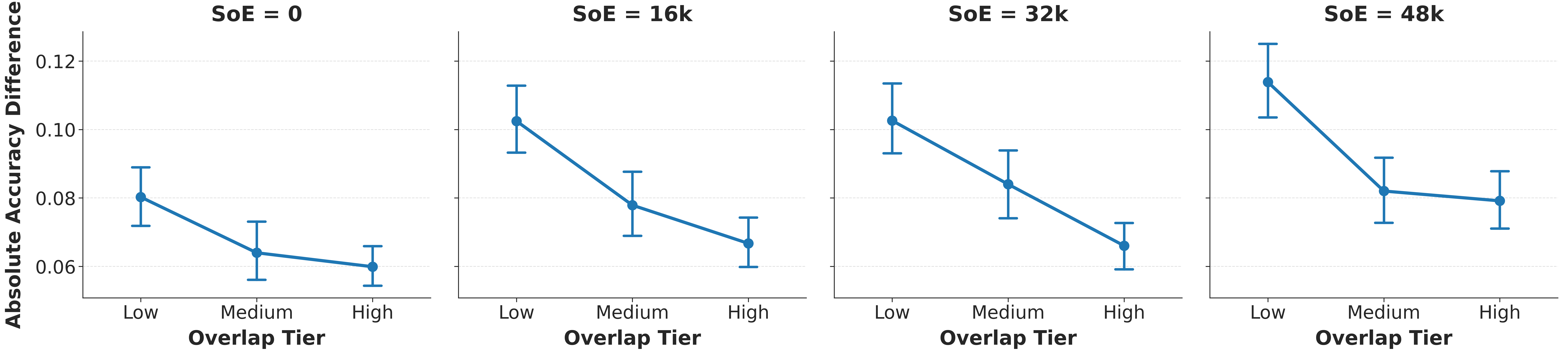}
    \caption{The absolute accuracy difference between primed and unprimed models relative to the degree of active neural overlap. Data points are binned into Low, Medium, and High overlap quantiles. A high degree of overlap consistently minimizes behavioral deviation, while low overlap leads to significant structural interference. This destabilizing effect of low overlap becomes increasingly severe at higher Step of Exposure (SoE) checkpoints.}
    \label{fig:diff_vs_overlap}
\end{figure*}

To further quantify this functional dependency, we examine how the degree of neural overlap dictates overall prediction stability. We calculate the absolute accuracy difference between the primed and unprimed models for each evaluated BLiMP category, a metric that captures the total behavioral deviation caused by the prime. To isolate underlying trends from the dense distribution of individual data points, we discretize the continuous shared neuron counts into three equal-sized quantiles (Low, Medium, and High Overlap) and visualize the mean absolute deviation alongside 95\% confidence intervals. Figure~\ref{fig:diff_vs_overlap} demonstrates a clear, persistent pattern. Categories that maintain a High overlap with the unprimed state exhibit significantly lower absolute accuracy differences, proving that successful activation of shared neurons stabilizes L2 performance. Conversely, Low overlap results in severe deviation. Crucially, the destabilizing effect of low overlap amplifies significantly as L1 dominance increases. At SoE=48k, the structural deviation in the Low overlap bin spikes dramatically. This dynamic perfectly aligns with our earlier hypothesis regarding L2 robustness. As the baseline L2 representations become sparser due to lower relative training exposure at higher SoE stages, failing to successfully recruit the shared neural workspace leaves the model entirely vulnerable to the dominant L1 structural prior, leading to massive grammatical interference.

\section{Discussion}

\subsection{Behavioral and Representational Evaluation}
We investigate the correlation between linguistic distance and L2 grammatical accuracy as evidence of crosslinguistic influence (CLI), aligning with previous approaches in the literature \cite{oba-etal-2023-second, yadavalli-etal-2023-slabert, constantinescu-etal-2025-investigating}. However, rather than analyzing raw accuracy, which can obscure subtle interactions, we decouple the analysis into distinct positive and negative transfer rates. In the baseline unprimed setting, we observe minimal variance across languages, demonstrating the capacity of language models to mask underlying transfer effects (Section \ref{sec:cli_distance}). Yet, as the Step of Exposure (SoE) increases, L1 entrenchment strengthens, yielding a highly significant correlation with linguistic distance ($\rho = -0.74$ and $\rho = -0.84$ for positive and negative transfer respectively, with $p < 0.05$ in both cases). Explicit priming significantly amplifies these transfer effects, mirroring human bilingualism where priming influences both grammatical and ungrammatical target structures \cite{hartsuiker2004, HARTSUIKER_BERNOLET_2017, vandijk2023, Baroncini_Torregrossa_2025}. As illustrated in Figure \ref{fig:prime_effect}, explicit priming significantly amplifies these transfer effects and their correlation with linguistic distance. Languages with shared syntactic structures successfully prime those structures in the L2, boosting positive transfer. Conversely, structurally divergent languages bias the model toward unacceptable predictions, increasing negative interference. Our layer-wise analysis (Section \ref{sec:layer_cli}) reveals that this priming effect does not manifest uniformly across the network. Instead, it emerges predominantly in the deeper layers. Beyond layer 6, a critical shift occurs where similar languages begin to drive higher positive and lower negative transfer, while distant languages diverge in the opposite direction. This delayed onset indicates that deeper layers are responsible for integrating the prime context into the final prediction. Furthermore, Section \ref{sec:dynamics_cli} demonstrates that this integration is grounded in structural typologies. The correlation between linguistic distance and the cosine similarity of surprisal differences strengthens and becomes statistically significant only in these deeper layers, perfectly tracking the behavioral transfer rates. Interestingly, in the unprimed condition, this correlation follows the exact opposite trajectory, moving from early significance to weaker, non-significant correlations in the final layers. The raw cosine similarity analysis (Figure \ref{fig:cosine_sim}) confirms that unprimed models ultimately converge toward highly similar, L1-agnostic L2 representations, though this convergence is hindered at higher SoE. This representational convergence suggests that the final layer of a transformer is not the optimal site for measuring CLI effects, potentially explaining the mixed findings regarding L1 transfer in prior literature \cite{aoyama-schneider-2024-modeling, constantinescu-etal-2025-investigating}. Broadly, these representational dynamics raise critical concerns for fine-tuning massive language models on low-resource languages. Because pretraining heavily entrenches specific linguistic structures, adapting to distant languages becomes increasingly difficult, leading to structural interference. This aligns with recent findings on plasticity loss \cite{hernandezgarcia2026scalesaveplasticityloss}, where models trained continuously on multiple languages struggle to adapt to held-out languages at later training stages.

\subsection{Mechanistic Effects}
Beyond behavioral evaluations, we investigate the mechanistic foundations of CLI. Neuroimaging research in human bilinguals demonstrates that the L1 and L2 activate overlapping brain regions \cite{RUSCHEMEYER2006354, consonni2012}, an overlap that is modulated by the linguistic distance between the two languages \cite{Tolentino_Tokowicz_2011}. To determine whether language models exhibit parallel architectural sharing, we identify and analyze syntactic L2 neurons. Section \ref{sec:neuron_distance} reveals that the cross-lingual overlap of these L2 neurons is modulated by linguistic distance, particularly after priming. This proves that the neural allocation for L2 grammar is not arbitrary but is physically shaped by pre-existing L1 structures. Moreover, the typological consistency of this L2 neuron allocation strengthens alongside L1 dominance. In the unprimed condition, the partial correlation between neuron overlap and linguistic distance shifts from being statistically insignificant at SoE=0 ($\rho = 0.12$, $p=0.225$) to significant at SoE=16k ($\rho = -0.23$, $p=0.021$). Under explicit priming, this correlation deepens continuously, reaching $\rho=-0.37$ at SoE=48k. This highlights a clear consequence of sequential training: prolonged exposure to the L1 deepens its structural entrenchment, which in turn exerts a stronger architectural effect on the neurons recruited for the L2. This sharing could indicate that the L2 actively leverages established L1 pathways, aligning with the shared-syntax hypothesis in psycholinguistics \cite{HARTSUIKER_BERNOLET_2017}. Crucially, the layer-wise distribution of these active neurons (Section \ref{sec:neuron_layer}) diverges sharply between the primed and unprimed states. Explicit priming forces the syntactic neurons to migrate toward the deeper layers where the prime context is actively integrated. This migration occurs uniformly across all languages, independent of their typological similarity to the L2. In human psycholinguistics, priming effects are traditionally attributed to the static activation of shared structural representations. The layer-wise migration observed in our models complicates a direct parallel, showing instead that the prime dynamically shifts the entire computational locus of L2 resolution. Despite the minimal absolute overlap between the unprimed and primed neural populations, Section \ref{sec:distance_overlap} demonstrates that the degree of shared neurons remains correlated with linguistic distance at higher SoE. Typologically similar L1s manage to activate a larger shared subspace with the unprimed state, whereas distant L1s force the model to rely on distinct neurons. Targeted ablations (Section \ref{sec:intervention_components}) confirm the functional importance of these overlapping neurons, as deactivating them degrades accuracy significantly more than ablating non-overlapping ones. Furthermore, under primed conditions, ablating Attention neurons yields a significantly sharper performance drop than ablating MLP units. This supports the hypothesis that crosslinguistic transfer is driven by the context integration capabilities of the attention mechanism. We propose that the observed neural migration occurs because early layers strictly compress the cross-lingual context, leaving the attention heads in the deeper layers to actively route that structural prior into the final prediction. Figure \ref{fig:delta_acc_distance} corroborates this mechanism, showing a negative correlation between ablation impact and linguistic distance specifically for Attention units. This confirms that the attention mechanism successfully leverages structural priors for similar languages while enforcing detrimental interference from distant ones, an effect that amplifies with increased L1 dominance.

\subsection{Role of L1 Dominance and L2 proficiency}
Throughout our analyses, L1 dominance emerges as the primary modulator of crosslinguistic influence (CLI), albeit with a critical tradeoff regarding L2 proficiency. Delaying the Step of Exposure (SoE) fundamentally entrenches L1 representations, which consistently increases the correlation between transfer metrics and linguistic distance across both behavioral (Section~\ref{sec:cli_effects}) and mechanistic evaluations (Section~\ref{sec:neuron_analysis}). These results parallel psycholinguistic studies demonstrating that CLI is jointly governed by L1 dominance and L2 proficiency \cite{HARTSUIKER_BERNOLET_2017, van_dijk_2022, Anthony_Ries_2025}. For instance, \citet{HARTSUIKER_BERNOLET_2017} posit that while low-proficiency L2 learners rely on transferring L1 syntactic information, L1-to-L2 crosslinguistic priming effects initially remain weak. These effects strengthen as L2 proficiency increases, which the authors attribute to the development of abstract, shared syntactic representations following sufficient L2 exposure. We observe similar dynamics in Section~\ref{sec:cli_effects}: as SoE increases---heightening L1 dominance while decreasing L2 proficiency---the linguistic consistency of L1 effects on the L2 strengthens for both positive and negative transfer. Crucially, however, positive transfer after priming (which relies on activating shared structures) eventually weakens at higher SoE stages due to diminished L2 proficiency. Conversely, negative transfer continues to intensify, as it does not require shared representations and is instead primarily driven by L1 dominance (e.g., total positive transfer---calculated as the sum difference between the primed and unprimed transfer---across languages rises from 1.032 at SoE=0 to 1.163 at SoE=16k, but drops to 1.140 at SoE=48k, whereas total negative transfer climbs continuously from 1.138 at SoE=0 to 1.597 at SoE=48k). Furthermore, our ablation of using linear learning rate scheduler (Section \ref{sec:ablate_linear}) which naturally weakens L2 proficiency at higher SoE checkpoints diminishes the positive priming effect while leaving the negative priming effect entirely intact. Our architectural evidence (Figure~\ref{fig:distance_vs_overlap}) corroborates this failure to develop shared representations; at higher SoE stages, the number of overlapping neurons between primed and unprimed states decreases, even for L1s that are highly similar to the L2. Furthermore, regarding the modulating effect of L1 dominance, \citet{van_dijk_2022} found that it predicts the strength, but not the presence, of CLI. Correspondingly, Section~\ref{sec:asymetry} shows that crosslinguistic priming in our models is strictly bidirectional, yet its magnitude is asymmetric, proving weaker when the target language is the dominant one. Ultimately, while this confirms a compelling parallel between artificial learners and psycholinguistic findings, drawing direct equivalencies should remain aspirational given the fundamental differences between human and artificial learners.

\section{Conclusion}

In this work, we presented a comprehensive behavioral and mechanistic investigation of Crosslinguistic Influence (CLI) in sequentially trained bilingual language models. By adapting psycholinguistic paradigms to artificial learners, we systematically varied the Step of Exposure (SoE) to control the interplay between first language (L1) dominance and second language (L2) proficiency. Utilizing crosslinguistic structural priming, we unmasked latent transfer dynamics, successfully decoupling the facilitation of grammatical structures (positive transfer) from the enforcement of ungrammatical interference (negative transfer). 

Our behavioral analyses demonstrated that early language entrenchment establishes a rigid structural prior that fundamentally dictates subsequent L2 acquisition. We found that increased L1 dominance strongly amplifies the correlation between linguistic distance and structural transfer. However, this dominance introduces a critical computational tradeoff. As prolonged L1 training diminishes relative L2 proficiency, the model loses its capacity to leverage shared structural representations for positive transfer, while negative interference from distant L1s continues to intensify. Furthermore, we established that while CLI is strictly bidirectional, its magnitude remains highly asymmetric and heavily dependent on the dominance status of the target language.

Beyond these behavioral observations, our mechanistic evaluations provided direct architectural evidence for how sequential training physically shapes parameter allocation. We showed that the cross-lingual overlap of L2 syntactic neurons is strictly modulated by linguistic distance, yielding a typological consistency that deepens as L1 entrenchment grows. Crucially, we uncovered that explicit structural priming induces a dynamic layer-wise neural migration, shifting the computational locus of grammatical resolution from the middle layers to the terminal layers. Through targeted causal ablations, we confirmed that deep-layer attention mechanisms are responsible for integrating this cross-lingual context, successfully routing the structural L1 prior into the final target prediction.

Ultimately, our findings confirm that crosslinguistic influence in artificial models is not an arbitrary artifact of capacity constraints. Instead, it is a highly structured, attention-driven phenomenon that mirrors the complex dynamics of L1 dominance and L2 proficiency observed in human bilinguals. These insights raise important considerations for the design of multilingual training curricula, highlighting the necessity of balancing exposure to prevent deeply entrenched dominant languages from enforcing detrimental structural interference on distant lower-resource targets. Future studies aiming to simulate second language acquisition using language models should consider systematically varying the Step of Exposure, evaluating CLI effects across intermediate layers rather than exclusively at the final layer, and potentially establishing a shared orthographic script between the L1 and L2 to more cleanly isolate underlying syntactic transfer.

\section{Limitations}

While our controlled experimental setup provides novel behavioral and mechanistic insights into crosslinguistic influence (CLI), several limitations should be considered when interpreting the results.

First, to enable the rigorous ablation of training dynamics (such as sequentially varying the Step of Exposure across 15 distinct language pairs), our experiments rely on training relatively small language models (GPT-2 architecture, 12 layers) from scratch. While this scale is standard for targeted mechanistic interpretability and computational psycholinguistics, it remains an open question whether the exact layer-wise neural migration and overlap dynamics scale proportionally to modern, massively parameterized large language models (LLMs) exhibiting emergent cross-lingual behaviors. 

Second, our behavioral evaluations are primarily constrained to the BLiMP and MultiBLiMP minimal pair benchmarks. While minimal pairs offer a highly controlled method for isolating specific syntactic phenomena like subject-verb agreement, they evaluate receptive grammatical preference rather than productive, free-form text generation. Consequently, our findings on structural transfer may not fully encapsulate discourse-level interference or the transfer of complex morphological agreements during open-ended generation tasks. 

Third, our crosslinguistic priming methodology relies on automatic translation (via NLLB) to generate the L1 primes. Although we explicitly ablated translation quality and demonstrated that the underlying typological distance rather than translation fidelity drives the observed transfer effects, machine translation can occasionally produce unnatural phrasing or ``translationese.'' Future work could benefit from using human-elicited primes to entirely remove any structural artifacts introduced by automated translation systems.

Fourth, while we conducted a reverse priming ablation to confirm the bidirectionality of CLI across 10 different target languages, the majority of our deep mechanistic analyses rely on English as the primary L2 target. Because English is a morphologically impoverished language with relatively strict word order, the specific mechanics of negative structural interference might manifest differently if the target L2 were highly agglutinative or featured free word order. 

Finally, we must emphasize the theoretical boundaries of this work. While psycholinguistic theories of human bilingualism served as the fundamental inspiration for our experimental design (e.g., decoupling positive and negative transfer, manipulating Step of Exposure, and utilizing structural priming), artificial neural networks are not cognitively equivalent to the human brain. The specific mechanisms we uncovered, such as the spatial migration of syntactic neurons and the reliance on deep-layer attention heads, are unique artifacts of the transformer architecture. 

\appendix
\section{Training Details}
\label{sec:training_details}

Our model training infrastructure builds upon the codebase provided by \citet{chang_word2022}\footnote{\url{https://github.com/tylerachang/word-acquisition-language-models}}. Because their original implementation natively supports only monolingual training, we extended the framework to accommodate bilingual training with flexible Step of Exposure (SoE) checkpoints and adjustable L1/L2 ratios. The training hyperparameters, detailed in Table~\ref{tab:hyperparams}, are largely adopted from \citet{chang_word2022} and \citet{arnett-etal-2025-acquisition}, though we introduce several critical modifications to suit our experimental design. 

Specifically, we increase the maximum sequence length from 128 to 256 tokens to ensure that the translation primes comfortably fit within the context window. To offset the memory overhead of this expansion, we proportionally halve the batch size from 128 to 64. Furthermore, we reduce the total training duration from the standard 128k steps down to 64k steps, enabling broader experimentation across 15 language pairs while maintaining a tractable computational cost. Finally, as motivated in Section~\ref{subsec:bilingual_training}, we strictly employ a constant learning rate scheduler rather than a standard linear decay to prevent confounding variables during SoE variation.

\begin{table}[h]
\scriptsize
\centering
\begin{tabular}{|l|c|}
\hline
\textbf{Hyperparameter} & \textbf{Value} \\ \hline
Layers & 12 \\
Embedding size & 768 \\
Hidden size & 768 \\
Intermediate hidden size & 3072 \\
Attention heads & 12 \\
Attention head size & 64 \\
Activation function & GELU \\
Vocab size & 50004 \\
Max sequence length & 256 \\
Position embedding & Absolute \\
Batch size & 64 \\
Train steps & 64k \\
Learning rate decay & Constant \\
Warmup steps & 5000 \\
Learning rate & $1 \times 10^{-4}$ \\
Adam $\epsilon$ & $1 \times 10^{-6}$ \\
Adam $\beta_1$ & 0.9 \\
Adam $\beta_2$ & 0.999 \\
Dropout & 0.1 \\
Attention dropout & 0.1 \\
Seed & 123 \\ \hline
\end{tabular}
\caption{Architectural and optimization hyperparameters for the GPT-2 bilingual models.}
\label{tab:hyperparams}
\end{table}

\section{Extended Analysis of Prime Translation Quality}
\label{sec:translation_appendix}

\begin{figure*}[ht!]
    \centering
    \includegraphics[width=1.\textwidth]{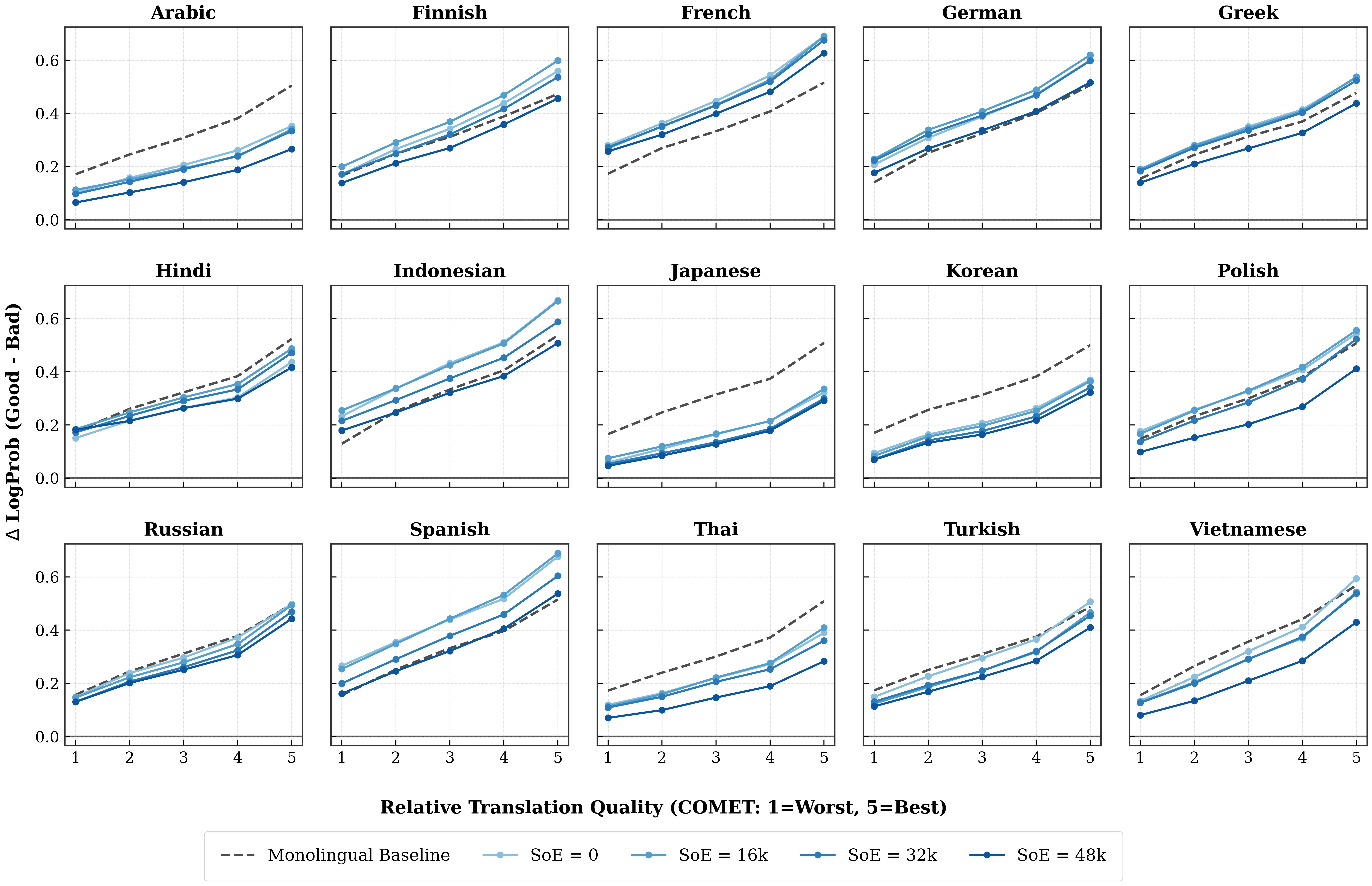}
    \caption{The effect of prime translation quality (measured via COMET) on $\Delta$ LogProb, broken down by individual L1 languages. While all languages exhibit a positive correlation between translation quality and grammatical preference, the vertical placement of the primed models depends on linguistic distance. Typologically similar languages generally sit above the monolingual baseline (positive transfer), whereas distant languages sit below it (negative transfer). The downward shift of the curves at higher SoE steps (darker lines) illustrates increased vulnerability to interference.}
    \label{fig:qe_per_language}
\end{figure*}

In Section~\ref{sec:translation_ablation}, we established the average relationship between prime translation quality and the $\Delta$ LogProb. However, aggregating these results obscures cross-lingual variations. Figure~\ref{fig:qe_per_language} disaggregates this data, presenting the individual trajectories for all 15 languages across the four Step of Exposure (SoE) checkpoints. 

The individual language plots corroborate our primary conclusion. The upward slope of the $\Delta$ LogProb as translation quality increases is universally consistent across all languages, tightly tracking the monolingual baseline. Crucially, the vertical offset of the primed trajectories relative to the baseline is dictated by typological proximity rather than translation quality. For structurally similar languages (e.g., French, German, and Spanish), the primed models consistently operate above the monolingual baseline, indicating robust positive transfer. Conversely, for distant languages (e.g., Arabic, Japanese, Korean, and Thai), the primed trajectories sit strictly below the baseline, manifesting as a persistent priming penalty. Furthermore, within almost every language panel, the trajectories corresponding to the highest SoE (Step 48k) consistently fall below the earlier checkpoints. This intra-language shift confirms that extended L1 entrenchment uniformly increases the model's vulnerability to structural interference.

\begin{figure*}[ht!]
    \centering
    \includegraphics[width=1.\textwidth]{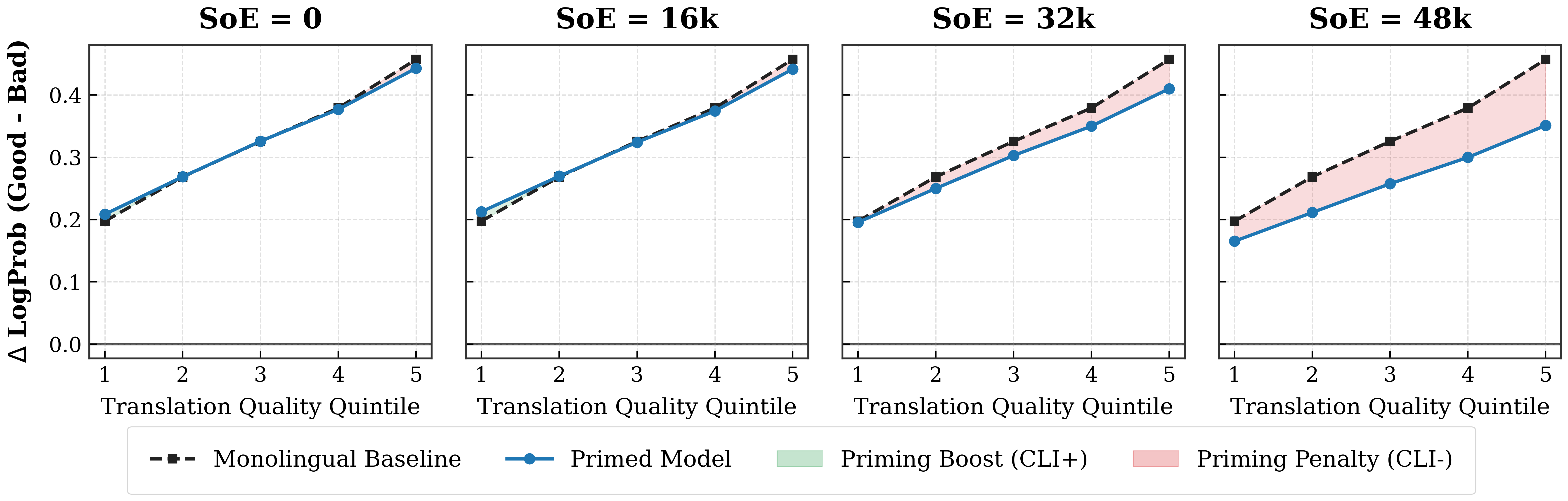}
    \caption{The aggregated effect of prime translation quality on $\Delta$ LogProb, measured using MetricX-XL. The consistent widening of the ``Priming Penalty'' (shaded red region) at higher translation qualities across the 32k and 48k checkpoints fully replicates the trends observed using COMET, confirming that highly idiomatic L1 primes induce greater structural interference.}
    \label{fig:qe_metricx}
\end{figure*}

To ensure these findings are not artifacts of the specific quality estimation model used (COMET), we replicate the aggregated analysis using MetricX-XL, an alternative state-of-the-art referenceless metric. As shown in Figure~\ref{fig:qe_metricx}, the MetricX-XL results perfectly mirror the prior COMET analysis. Both the primed and monolingual models show improved grammatical preference with better translation scores, yet a widening priming penalty emerges at higher SoE stages (32k and 48k) as translation quality increases. 

Together, these granular and cross-metric analyses definitively confirm that translation quality does not arbitrarily drive crosslinguistic influence. Instead, the magnitude and direction of structural transfer are fundamentally modulated by linguistic distance and the relative degree of L1 dominance.

\begin{acknowledgments}

\end{acknowledgments}

\bibliographystyle{compling}
\bibliography{COLI_template}

\end{document}